%% file: _main.tex
\documentclass[manuscript,nonacm]{acmart}

\AtBeginDocument{%
  }

\input{Packages.tex}

\usepackage{xcolor}

\begin{document}

\title{\protect\input{1_Content/app_name} - Pattern-based Risk Estimation Of Remaining Explosives}

\author{Björn Kischelewski}
\authornote{Equal contribution.}
\email{bjoern@kischelewski.de}
\orcid{0009-0008-3673-4571}
\affiliation{%
  \institution{AI Centre, Department of Computer Science, University College London}
  \city{London}
  \country{United Kingdom}
}
\author{Benjamin Guedj}
\authornotemark[1]
\email{b.guedj@ucl.ac.uk}
\orcid{0000-0003-1237-7430}
\affiliation{%
  \institution{AI Centre, Department of Computer Science, University College London}
  \city{London}
  \country{United Kingdom}
}
\affiliation{%
  \institution{Inria}
  \country{France}
}
\author{David Wahl}
\email{david.n.wahl@gmail.com}
\orcid{0009-0007-8627-2638}
\affiliation{%
  \institution{Scientia Education}
  \country{Switzerland}
}

\begin{abstract}
  \input{1_Content/100_Abstract}
\end{abstract}

\maketitle

\clearpage

\input{1_Content/110_Introduction}

\input{1_Content/120_Related_Work}

\input{1_Content/130_Methodology}

\input{1_Content/140_Experiments}

\input{1_Content/150_Results}

\input{1_Content/160_Discussion}

\input{1_Content/170_Conclusion}

\begin{acks}
\input{1_Content/190_Acknowledgements}
\end{acks}

\bibliographystyle{Reference-Format}
\bibliography{paperpile_pub}

\end{document}

%% file: Packages.tex
\usepackage{cleveref}

\usepackage{algorithm}
\usepackage{algpseudocode}

\usepackage{graphics} 

\usepackage{tabularx}
\newcolumntype{Z}{>{\raggedright\arraybackslash}X}

\usepackage{subcaption}

\usepackage{multicol}

%% file: 1_Content/app_name.tex
\textsc{RestoreAI}\unskip

%% file: 1_Content/100_Abstract.tex
Today, over 60 countries are affected by landmines, with more than 4,700 people killed or injured in 2022. However, landmine removal remains a slow and resource-intensive process. Clearance operations cover vast areas, while skilled deminers, hardware, and funding are limited.
To address this challenge, artificial intelligence (AI) tools have been proposed to improve the detection of explosive ordnance (EO). Existing approaches have focused primarily on landmine recognition, with limited attention to prediction of landmine risk based on spatial pattern information.

This work aims to answer the following research question: How can AI be used to predict landmine risk from landmine patterns to improve clearance time efficiency? 

To that effect, we introduce \textbf{\input{1_Content/app_name}}, an AI system for pattern-based risk estimation of remaining explosives. 
\input{1_Content/app_name} is the first AI system that leverages landmine patterns for risk prediction, improving the accuracy of estimating the residual risk of missing EO prior to land release. 
We particularly focus on the implementation of three instances of \input{1_Content/app_name}, respectively, linear, curved and Bayesian pattern deminers. First, the linear pattern deminer uses linear landmine patterns from a principal component analysis (PCA) for the landmine risk prediction. Second, the curved pattern deminer uses curved landmine patterns from principal curves. Finally, the Bayesian pattern deminer incorporates prior expert knowledge by using a Bayesian pattern risk prediction. 

Performance evaluation on real-world landmine data shows a significant increase in clearance time efficiency by incorporating landmine pattern information into the prediction. In particular, the best pattern-based deminers achieve a 14.37 percentage points higher average share of cleared landmines per timestep and require 24.45\% less time than the best baseline deminer to find all landmines. Interestingly, no significant performance differences are found between the linear and curved pattern deminers, suggesting the merits of using simpler and more efficient linear patterns for the risk prediction, lowering the computational burden of our algorithm when used in the field. \input{1_Content/app_name} has the potential to improve the safety of the skilled deminers, and the efficiency and agility of the entire demining process.

%% file: 1_Content/110_Introduction.tex
\section{Introduction} \label{Chap1}

The \citet{International-Campaign-to-Ban-Landmines2023-pc} reports in its Landmine Monitor 2023 that 60 countries suffer from landmines and other explosive remnants of war (ERW), and more than 4,700 people were killed or injured by these explosives in 2022. Of these casualties, 85\% are civilians, including more than 1,000 children. In addition, ongoing conflicts such as those in Ukraine, Gaza and Yemen, are leading to new uses of explosive ordnance (EO) \citep{Cumming-Bruce2023-ry}. For example, Ukraine was the country with the second highest number of EO casualties in the world in 2022, a tenfold increase from 2021 \citep{International-Campaign-to-Ban-Landmines2023-pc}. 
Efficient and effective clearance of affected areas will therefore save lives. In addition, other positive effects of EO clearance are identified by \citet{Hofmann2017-pl}. These effects are found to be directly linked to 12 of the 17 Sustainable Development Goals (SDGs) and indirectly linked to four other SDGs, highlighting the high social, environmental and economic impact of EO clearance operations.
However, despite increasing efforts to identify, clear and release affected regions, most countries that have signed the Oslo Treaty on a global ban on anti-personnel (AP) landmines are likely to fall short of the goal of clearing all landmines by 2025 
\citep{Cumming-Bruce2023-ry, International-Campaign-to-Ban-Landmines2023-pc, Anti-Personnel-Mine-Ban-Convention2019-zy}. 
Ukraine alone has reported an area of at least 160,000km$^2$ requiring  survey and has documented landmines in 11 of its 27 regions \citep{International-Campaign-to-Ban-Landmines2023-pc, ACAPS2024-xy}. 
A GLOBSEC report by \citet{Osmolovska2023-ze} estimates that it would take more than 750 years to fully clear this area using current resources. 

At the same time, landmine clearance is a challenging process for three main reasons. 
First, the size of hazardous areas is often overestimated and deminers have difficulty distinguishing between a suspected and confirmed hazardous area \citep{Mine-Action-Review2023-it, Cumming-Bruce2023-ry}. 
Second, the average deminer can only clear about 60 square meters per day and has even lower productivity in difficult terrain \citep{Harutyunyan2023-qc, Geneva-International-Centre-for-Humanitarian-Demining2023-om}. 
Third, landmine clearance operations usually lack funding and skilled deminers \citep{ACAPS2024-xy}. 
As a consequence of these challenges, \citet{Harutyunyan2023-qc} find that the median time spent per deminer to find and clear a landmine is 67.5 days, with some clearance operations reporting that a team has spent more than a year finding each landmine.

Therefore, several public organizations and NGOs call for innovative research in technology to improve the efficiency of landmine clearance \citep{Anti-Personnel-Mine-Ban-Convention2019-zy, United-Nations-General-Assembly2023-wn, Harutyunyan2023-qc, Geneva-International-Centre-for-Humanitarian-Demining2024-ya, Federal-Foreign-Office-Germany2024-sl}. 
For example, the Oslo Action Plan from the \citet{Anti-Personnel-Mine-Ban-Convention2019-zy} explicitly states in Action 27 to "[t]ake appropriate steps to improve the effectiveness and efficiency of survey and clearance, including by promoting the research, application and sharing of innovative technological means to this effect". 

In particular, innovations in information management, data analytics and AI will play a key role in improving the efficiency of landmine clearance \citep{Osmolovska2024-mi, United-Nations2023-zs, Toscano2021-im}.
In response, a significant amount of research has been conducted on the application of AI to landmine clearance \citep{Kischelewski2025-qy}. However, this research focuses primarily on AI for landmine object detection from sensor data, such as ground penetrating radar (GPR) or metal detector responses. In contrast, only few publications investigate AI for landmine risk prediction. Such risk prediction can contribute to a more effective resource allocation and thus, increase the efficiency of landmine clearance. In particular, there is no research on AI for landmine risk prediction using knowledge of landmine patterns \citep{Kischelewski2025-qy}.

Therefore, we investigate the following research question:
\begin{quote}
\begin{center}
\textbf{How can AI algorithms be used for landmine risk prediction from landmine patterns to improve landmine clearance time efficiency?}
\end{center}
\end{quote}
In order to answer this research question, we introduce \textbf{\input{1_Content/app_name}}, an AI system for risk estimation of remaining explosives. The solution comprises a multi-step machine learning (ML) model that predicts landmine risk from patterns of found landmines for a region of interest. In particular, three instances of \input{1_Content/app_name} with different approaches to landmine risk prediction are compared to provide a robust answer to the research question.
\begin{itemize}
    \item \textbf{Linear Pattern Deminer:} A virtual deminer which uses linear landmine pattern information to predict landmine risk and update its demining route online.
    \item \textbf{Curved Pattern Deminer:} A virtual deminer which uses curved landmine patterns information to predict landmine risk and update its demining route online.
    \item \textbf{Bayesian Pattern Deminer:} A virtual deminer which uses curved landmine patterns information with prior expert knowledge to predict landmine risk and update its demining route online.
\end{itemize}
Finally, the clearance performance of all three instances of \input{1_Content/app_name} is compared to the performance of baseline deminers. For this performance evaluation, a virtual demining environment is created using real-world landmine test data, and the \textbf{Demining Score} is introduced to measure clearance performance as the average share of landmines found over the entire clearance process (see \Cref{Sec_Evaluation} for more details on performance evaluation).

The present article is organized into four sections. 
\Cref{Chap2} provides a comprehensive overview of the research on AI for landmine risk prediction and current gaps in research.
\Cref{Chap3} introduces the \input{1_Content/app_name} system and its underlying statistical methods and ML techniques.
\Cref{Chap4} details the real-world dataset and experimental setup used to train, evaluate and compare the three instances of \input{1_Content/app_name}. 
\Cref{Chap5} presents the performance results of testing the instances on landmine clearance data and compares them to the performance of baseline models.
Finally, \Cref{Chap6} discusses the implications of the results for the research question and for real-world landmine clearance operations, as well as its limitations and ethical considerations. We gather concluding thoughts in \Cref{Chap7}.

%% file: 1_Content/120_Related_Work.tex
\section{Related Work} \label{Chap2}

Our previous work contributed an analysis of the related work on AI systems for EO detection in clearance operations in a comprehensive literature review \citep{Kischelewski2025-qy}. We found that research on AI systems for EO detection in clearance operation can be grouped into two main streams. 
The majority of research focuses on EO object detection, including AI systems to improve EO detection from live sensor data. In contrast, a minority of research focuses on EO risk prediction. These AI systems use EO-related and non-EO-related geographic input data to predict the risk of finding EO at a specific location in a region of interest. Interestingly, most publications on EO risk prediction use a combination of EO-related and non-EO-related data. 

The non-EO-related data includes various information about the topology and infrastructure of the region of interest. Topological features include information on elevation, incline, land use, forests, rivers, animal density, soil texture, temperature, precipitation, and visibility \citep{Riese2006-bm, Rafique2019-pt, Rubio2023-hy, Saliba2024-lx, Bajic2010-mu}. In addition, infrastructure features include information on roads, railways, airfields, seaports, bridges, cities, buildings, financial institutions, schools, borders, telecommunication lines, power lines, oil lines, orchards, bunkers, trenches, and shelters \citep{Riese2006-bm, Rafique2019-pt, Rubio2023-hy, Saliba2024-lx, Bajic2010-mu}. 

The EO-related data includes information from EO records and data from EO incidents in the region of interest. For example, \citet{Alegria2011-uq} use only landmine incident data to train their model. \citet{Riese2006-bm} also use several EO-related features, such as the distance to the nearest minefield, the distance to the confrontation line, and the distance to the nearest recorded EO incident, to train their model. 
However, none of these publications leverages the fact that EO, in particular landmines, are usually not placed randomly but follow a pattern. \citet{Thomas2010-vg} show that such patterns can be used to improve landmine detection performance from sensor data. Thus, it is likely that pattern information can also improve landmine risk prediction.

%% file: 1_Content/130_Methodology.tex
\section{Methodology} \label{Chap3}

\begin{figure}[H]
    \centering
    \includegraphics[width=\textwidth]{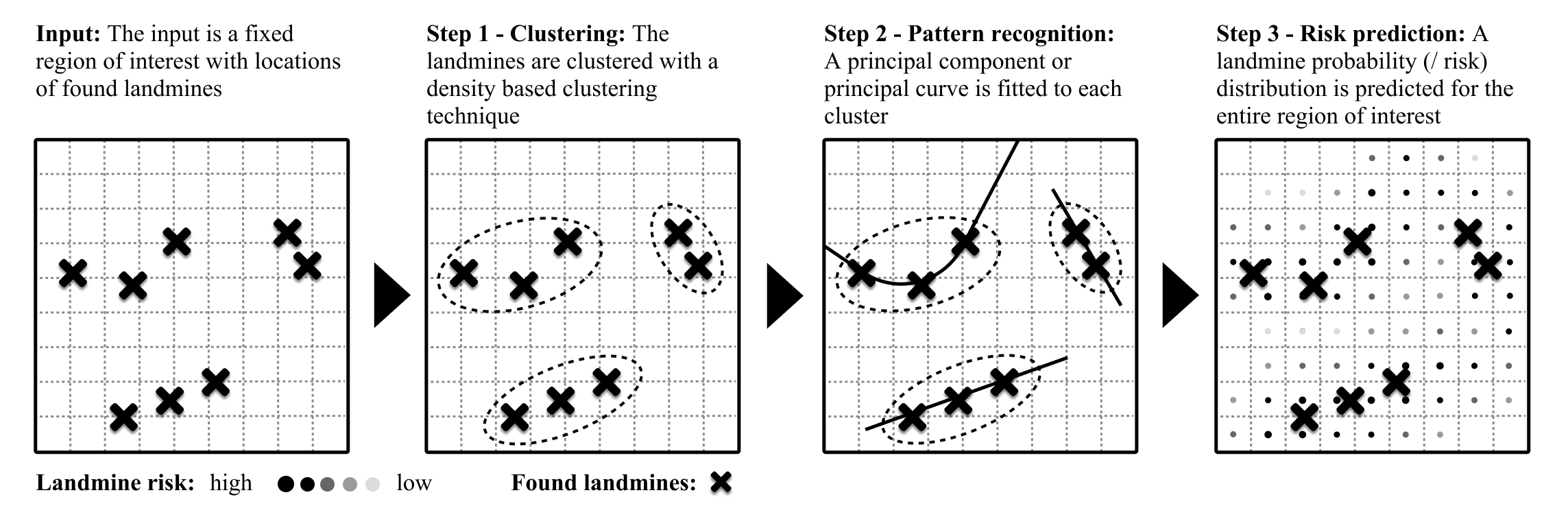}
    \caption{\protect\input{1_Content/app_name} pattern-based landmine risk prediction concept}
    \label{fig:13_Pattern_Concept}
\end{figure}

We introduce \input{1_Content/app_name}, an AI system for risk estimation of remaining explosives. The solution is a multi-step ML model that predicts landmine risk for a region of interest based on patterns of found landmines. The pattern-based landmine risk prediction is a three-step process (see \Cref{fig:13_Pattern_Concept}). The prediction works with information about the locations of landmines found in the region of interest. 
First, the landmines are clustered using the density-based clustering technique DBSCAN. 
Second, the patterns for each cluster are identified using PCA or principal curves. 
Third, the landmine risk is predicted from each landmine pattern for a region of interest using (Bayesian) logistic regression. 
Finally, multiple risks from different landmine patterns are combined to obtain a total pattern-based landmine risk for a region of interest. 

Further, the problem of landmine risk prediction for a region of interested is abstracted by applying a grid structure to this region. The grid spans regular, non-overlapping square tiles over the entire region. This approach reduces the problem to predicting a single landmine risk for each tile. Consequently, the tile size affects the effectiveness (granularity) and efficiency (computational time) of the prediction. The tiling is set to 25 by 25 meters because this area could be cleared in a reasonable amount of time. For example, a team of ten deminers would need about one day on average to clear a 25 by 25 meters tile, assuming a median clearance rate of 60 square meters per day per deminer as reported by \citet{Harutyunyan2023-qc}. 

The proposed approach also allows the landmine risk to be updated online during clearance operations. This is achieved by updating the pattern-based risk prediction each time a new landmine is found. This approach aligns well with real-world clearance operations, which gain information about found landmines during the process. 

Overall, three instances of \input{1_Content/app_name} are trained, tested and compared: 
\begin{itemize}
    \item \textbf{Linear Pattern Deminer:} A virtual deminer which uses linear landmine pattern information to predict landmine risk and update its demining route online.
    \item \textbf{Curved Pattern Deminer:} A virtual deminer which uses curved landmine patterns information to predict landmine risk and update its demining route online.
    \item \textbf{Bayesian Pattern Deminer:} A virtual deminer which uses curved landmine patterns information with prior expert knowledge to predict landmine risk and update its demining route online.
\end{itemize}
\Cref{table:13_Versions_implementation} provides a detailed overview of the underlying technical implementation of each instance at each step of the process. 

\begin{center}
\begin{table}[H]
\smaller
\begin{tabularx}{\textwidth}{|lZZZ|} 
 \hline
 \textbf{Instance} & \textbf{Clustering} & \textbf{Pattern recognition} & \textbf{Pattern risk prediction}  \\ [0.5ex] 
 \hline
 Linear Pattern Deminer & DBSCAN & PCA & Logistic regression \\ 
 \hline
 Curved Pattern Deminer & DBSCAN & Principal curves & Logistic regression \\ 
 \hline
 Bayesian Pattern Deminer & DBSCAN & Principal curves & Bayesian logistic regression \\ 
\hline
\end{tabularx}
\caption{Implementation details for each instance of \protect\input{1_Content/app_name}}
\label{table:13_Versions_implementation}
\end{table}
\end{center}

\subsection{Clustering}

To identify patterns, the samples of found landmines are first clustered using the density-based clustering algorithm DBSCAN introduced by \citet{Ester1996-yc}. The algorithm defines a cluster so "that for each point of a cluster the neighborhood of a given radius has to contain at least a minimum number of points, \emph{i.e.}, the density in the neighborhood has to exceed some threshold" \citep{Ester1996-yc}. Thus, the algorithm allows for clusters of arbitrary shape and is well-suited to cluster landmines which can occur in various non-spherical patterns. 

\begin{figure}[H]
    \centering
    \includegraphics[width=200pt]{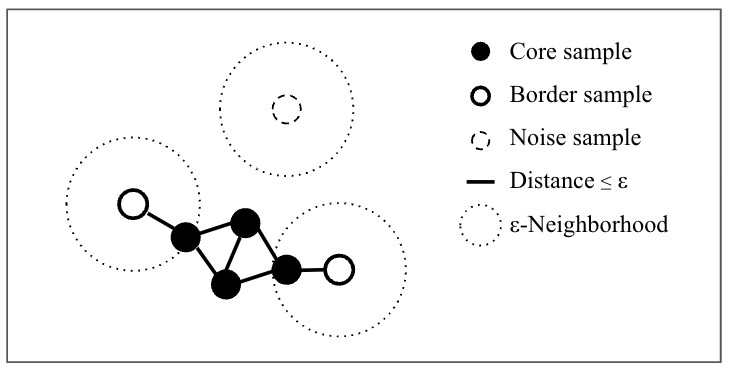}
    \caption{Exemplary DBSCAN cluster assignment}
    \label{fig:12_DBSCAN}
\end{figure}

DBSCAN defines the neighborhood distance $\epsilon$ and the minimum number of samples in this neighborhood $\mathrm{MinPts}$ globally. 
Then, the $\epsilon$-neighborhood of a sample $p \in X$ is defined as 
\begin{equation*}
N_{\epsilon}(p) = \{q \in X \, | \, \mathrm{distance}(p,q) \leq \epsilon \} \;. 
\end{equation*}
Furthermore, \citet{Ester1996-yc} differentiate between core and border samples of a cluster. 
A core sample $q$ has at least $\mathrm{MinPts}$ in its $\epsilon$-neighborhood, so that $N_{\epsilon}(q) \geq \mathrm{MinPts}$. 
A border sample $p$ only has to be \emph{directly density-reachable}, meaning it has to be in the $\epsilon$-neighborhood of a core sample $q$, so that $p\in N_{\epsilon}(q)$.
In addition, \citet{Ester1996-yc} define that a point $p_n$ is \emph{density-reachable} from a point $p_1$ if a chain of points $p_1, p_2, ..., p_{n-1}, p_n$ exists such that $p_{i+1}$ is directly density-reachable from $p_i$, for all $i\in\{1, ..., n-1\}$.
In practice, the algorithm builds a cluster by starting with an arbitrary core sample and adding all samples in its $\epsilon$-neighborhood to the cluster. If one of the added samples is itself a core sample, the process is repeated for that sample \citep{Schubert2017-pv}. All samples that are not \emph{density-reachable} from any core sample are considered noise \citep{Ester1996-yc, Schubert2017-pv}. \Cref{fig:12_DBSCAN} shows the concept of DBSCAN for an exemplary dataset with $\mathrm{MinPts} = 2$.

We use the \texttt{scikit-learn} implementation of DBSCAN by \citet{Pedregosa2011-nt}. The Euclidean distance is used as proximity metric, because it represents the actual physical distance in the metric space. Also, the $\mathrm{MinPts}$ parameter is set to one because landmines often form a linear or curved pattern with equal distances between them. The $\epsilon$ value is used as hyperparameter \emph{cluster\_max\_distance} of the model (compare \Cref{table:13_Hyperparameters}). Finally, the algorithm yields $m$ clusters, each of the form $C=\{c_1, c_2, ..., c_n\}$ with $c'$ defined as the center point of the cluster.

\subsection{Pattern Recognition}
Next, a pattern is identified for each cluster by performing a PCA or fitting a principal curve. 

\subsubsection{Linear Pattern Recognition}
A two-dimensional PCA is used by the Linear Pattern Deminer to identify a linear pattern for a cluster $C$. PCA is an unsupervised ML technique that transforms the data "to a new set of variables, the principal components, which are uncorrelated, and which are ordered so that the first few retain most of the variation present in all of the original variables" \citep{Jolliffe2002-hq}. In a two-dimensional space, PCA dimensionality reduction can be used to detect straight lines by extracting the first principal component from a set of data points \citep{Lee2006-qe}. For an independent and identically distributed dataset, a principal component can be computed either by maximizing the variance of the projected data or by minimizing the average projection costs of the data as described by \citet{Bishop2006-pl}.  

We use the PCA implementation of \texttt{scikit-learn} by \citet{Pedregosa2011-nt}. In particular, the first principal component, \emph{i.e.}, the principal component that captures the highest variance of the cluster, is used as the linear pattern. The \texttt{scikit-learn} package provides a \emph{transform}-function to represent each point $p$ in terms of the first principal component. If $p'$ is the orthogonal projection of $p$ onto the first principal component in the original space, then the new representation of $p$ after transformation is defined by the \textbf{distance} to the principal component
\begin{equation*}
\delta = \left\Vert p' - p \right\Vert^2 \;,
\end{equation*}
and the \textbf{progress} 
\begin{equation*}
\gamma = \left\Vert p' - c' \right\Vert^2 \;,
\end{equation*}
which is the distance of the projection to the cluster center on the principal component.

\subsubsection{Curved Pattern Recognition}
The Curved Pattern Deminer and the Bayesian Pattern Deminer both identify a curved pattern for a cluster $C$ by fitting a principal curve to the cluster. 
A principal curve can be interpreted as a nonlinear principal component of a dataset. It was first introduced by \citet{Hastie1989-wt}.
The authors define a one-dimensional principal curve for a random variable $\textbf{X}$ in a $d$-dimensional space as a vector $\textbf{f}(\lambda)$ of $d$ coordinate functions. Each coordinate function is parameterized by the variable $\lambda$ which provides an ordering along the principal curve. 
Further, they define the projection index $\lambda_{\textbf{f}}(x)$ for all $x \in \textbf{X}$ as the largest $\lambda$ value of $\textbf{f}$ for which the projection of $x$ onto the principal curve has minimum distance, such that
\begin{equation*}
\lambda_{\textbf{f}}(x) = \sup_{\lambda}\ \{\lambda:||x - \textbf{f}(\lambda)|| = \inf_{\mu}||x - \textbf{f}(\mu)|| \} \;.
\end{equation*}
The vector $\textbf{f}(\lambda)$ is a principal curve if it is self-consistent, \textit{i.e.}, if each $\textbf{f}(\lambda)$ is the mean of all points that project onto it. Hence,
\begin{equation*}
\begin{split}
&\textbf{f}(\lambda) \text{ is a principal curve}    \\
\Leftrightarrow \;  &\textbf{f}(\lambda) \text{ is self-consistent}    \\
\Leftrightarrow \; &\forall \lambda: \textbf{f}(\lambda) = E(\textbf{X} \, | \, \lambda_{\textbf{f}}(x) = \lambda) \;.
\end{split}
\end{equation*}

\citet{Hastie1989-wt} propose the following iterative algorithm to calculate a principal curve: 
\begin{algorithm}[H]
\caption{Principal curve algorithm by Hastie and Stuetzle}
\label{alg:PrincipalCurve}
\begin{algorithmic}[1]
\State Initialize $\textbf{f}^{(0)}(\lambda)$ as the first principal component.
\Repeat
    \State Set $\textbf{f}^{(j)}(\lambda) =  E(\textbf{X} | \lambda_{f^{(j-1)}}(x) = \lambda) \; , \; \forall \lambda$.
    \State Define $\lambda^{(j)}(x) = \lambda_{f^{(j)}}(x) \; , \; \forall x \in \textbf{X}$.
    \State Transform $\lambda^{(j)}$ so that $\textbf{f}^{(j)}$ is unit speed.
    \State Evaluate $D(X, \textbf{f}^{(j)}) = E_{\lambda^{(j)}} [ ||\textbf{X} - \textbf{f}(\lambda^{(j)}(\textbf{X}))||^2 \, | \, \lambda^{(j)}(\textbf{X})] $.
\Until{the change in $D(X, \textbf{f}^{(j)})$ is below some threshold $\epsilon$}.
\end{algorithmic}
\end{algorithm}

However, this version of the algorithm only works for data from a random variable \textbf{X} and must be adapted to work with a finite dataset $C=\{c_1, c_2, ..., c_n\}$, the set of landmine coordinates in a cluster. 
For this, \citet{Hastie1989-wt} represent the principal curve $\textbf{f}(\lambda)$ by $n$ tuples $(\lambda_i, \textbf{f}_i)$, each defining a polygon, and define $\lambda_i$ as the arc length from $\textbf{f}_1$ to $\textbf{f}_i$ with $\lambda_1 = 0$.
Furthermore, the projection step is adapted to the new definition of $\textbf{f}(\lambda)$. The projection index $\lambda_{\textbf{f}}(x)$ is defined as the projection index $\lambda_{\textbf{f}_i}(x)$ of the segment $\textbf{f}_i$ for which the projection distance of $x$ onto the segment is smallest.
Also, the conditional expectation step is adjusted for a finite dataset $C$ because often only one sample projects onto $\textbf{f}$ at a specific $\lambda$ value. Therefore, a smoothing spline can be used to fit the principal curve. As described by  \citet{Hastie1989-wt}, such a spline averages multiple samples $c_k$ for which $\lambda_{\textbf{f}}(c_k)$ is in a defined range around the specific $\lambda$ value. Although this implementation of the algorithm is not proven to converge, it performs well in experiments.

In particular, the \texttt{pcurvepy} Python package from \citet{Zhang2020-ks} is implemented with slight modifications to control the smoothness factor for the smoothing spline.  
A maximum degree of two is used for the spline polynomials to avoid unrealistic, highly complex patterns. However, if only two landmines are part of the cluster, degree one polynomials are used for the spline. Also, the smoothness factor for the smoothing spline is used as a hyperparameter \emph{pc\_smoothness\_factor} of the model (see \Cref{table:13_Hyperparameters}). 

\begin{figure}[h]
   \centering
   \includegraphics[width=300pt]{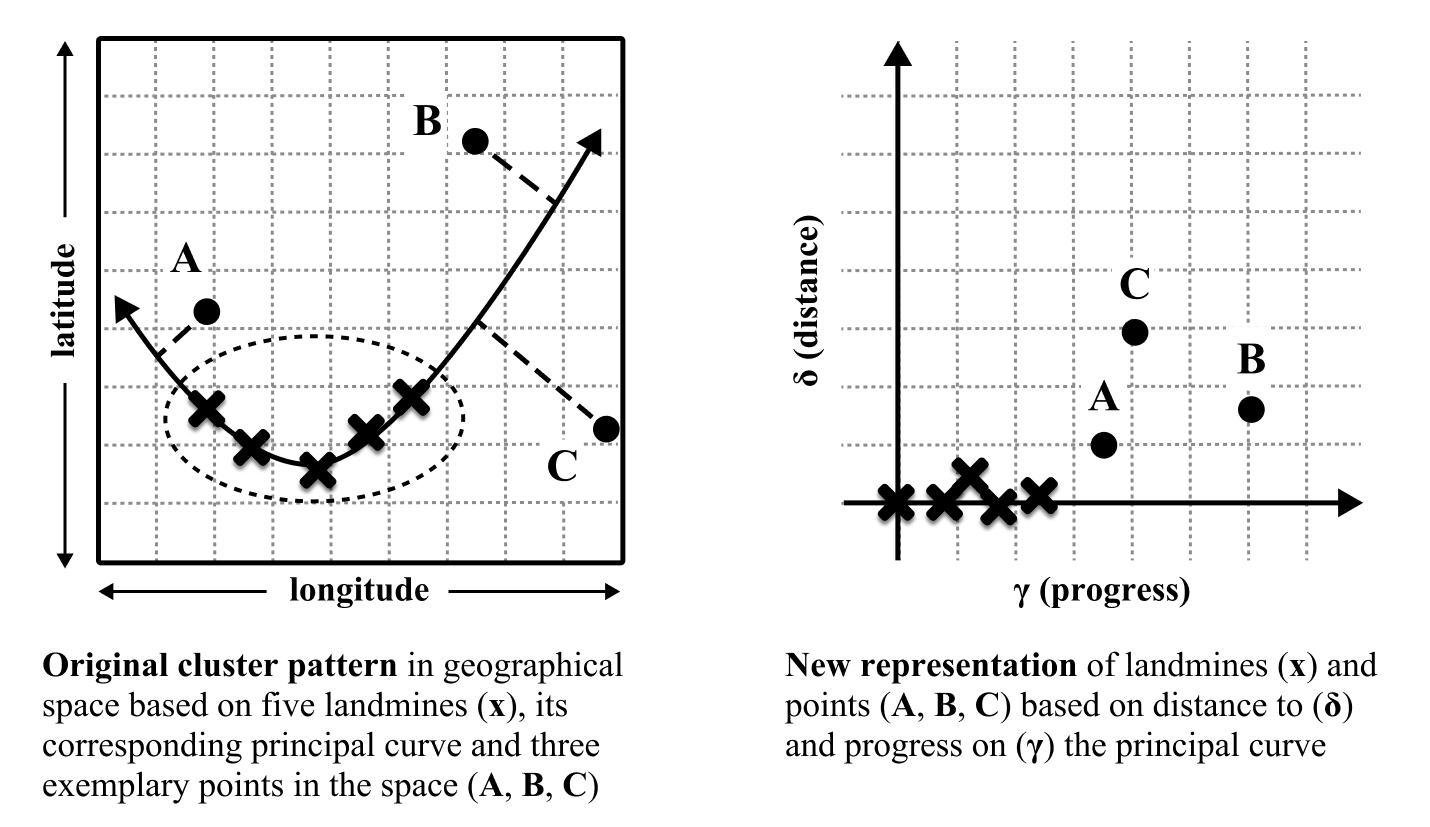}
   \caption{Exemplary change of representation for a principal curve}
   \label{fig:13_Representation}
\end{figure}

Predicting the risk from the curved pattern for each point in the region of interest requires a change in representation of the space similar to the transformation of the PCA. For a cluster $C$ and its respective principal curve $\textbf{f}(\lambda)$, the projection $p'$ of a point $p$ is defined as the point on $\textbf{f}(\lambda)$ with minimal distance to $p$. Hence, each point $p$ can be expressed by $\delta$, the Euclidean \textbf{distance} from $p$ to $p'$, and $\gamma$, the \textbf{progress} of $p'$ on the principal curve, which is the arc distance from $p'$ to the cluster center on the arc of the principal curve. \Cref{fig:13_Representation} shows an example of the change in representation for a cluster of five landmines, its respective principal curve and three exemplary points in the geographic space. 

However, the exact calculation of distance and progress for a point requires the calculation of the minimum distance to a spline and the calculation of the arc length of a spline which do not have closed-form solutions. Therefore, an approximation is used to calculate the distance and the progress. 
In detail, each principal curve is represented by a set of approximation points $P_{\mathrm{sp}} = \{s_1, s_2, ..., s_n \}$ on the principal curve, which cover an approximation distance $d_{\mathrm{approx}}$ in both directions from the center of the curve. This approximation distance is set to 1,000 meters for training. After training, the distance is set to the minimum progress for which the system predicts a pattern-based landmine risk of less than 5\% at a distance value of zero. 
In addition, the distance between the approximation points is set to 25 meters to match the tile size of the grids and allow for efficient computation.
Finally, the projection of a point $p$ onto the principal curve is approximated as $\tilde{p}' \in P_{\mathrm{sp}}$, such that
\begin{equation*}
\tilde{p}' \; = \text{arg}\min_{s} \; \left\Vert s - p \right\Vert^2 \;,\; s \in P_{\mathrm{sp}} \;,
\end{equation*}
and the set of approximation points from the cluster center to $\tilde{p}'$ is defined as $\tilde{Z}_{} = \{s_1, s_2, ..., \tilde{p}' \}\,\subset\,P_{\mathrm{sp}} \;$.
Therefore, distance and progress of $p$ can be approximated as 
\begin{equation*}
\begin{split}
\tilde{\delta} \; &= \left\Vert \tilde{p}' - p \right\Vert^2 
\\
\tilde{\gamma} \; &= \sum_{i=1}^{|Z|-1} \left\Vert \tilde{Z}_{i} - \tilde{Z}_{i+1} \right\Vert^2 \;.
\end{split}
\end{equation*}

\subsection{Risk Prediction}
These new representations allow to fit a classifier that predicts the landmine probability for a point in the region of interest based on a pattern. 
For each landmine cluster in the training regions, the respective PCA or principal curve is fitted and used to transform all landmine locations and all center points of tiles without any landmine to the new representation. The target value to learn is set to one for all landmine coordinates, and to zero for all center points of tiles without any landmine. 
Furthermore, the weight of the landmine coordinates in the training data is used as hyperparameter of the model (\emph{landmine\_weight}; see \Cref{table:13_Hyperparameters}). A higher weight of these positive samples is beneficial for the model to work against the under representation of landmines in the training data.

\subsubsection{Frequentist Prediction} 
For the Linear Pattern Deminer and the Curved Pattern Deminer, the training data is scaled using a standard scaler and a logistic regression classifier is fit to the data.
The logistic regression classifier assumes a linear relation 
\begin{equation*}
z_i = \beta_0 + \beta^T x_i
\end{equation*}
between the target variable and an input feature $x_i$ \citep{McCullagh1989-bx, Hastie2009-ud}.
To ensure a valid probability, the sigmoid function is applied to this linear relation \citep{Jurafsky2008-wu, McCullagh1989-bx, Hastie2009-ud}. Hence, the probability of a sample $x_i$ belonging to the positive class, \textit{i.e.} containing a landmine, is
\begin{equation*}
p(x_i = 1) = \frac{1}{1+ e^{-z_i}} = \frac{1}{1+ e^{-(\beta_0 + \beta^T x_i)}} \;.
\end{equation*}
The logistic regression is fitted using maximum likelihood estimation \citep{Hastie2009-ud, Shalev-Shwartz2014-tc}. 
No regularization is applied because the weights (for the features $\delta$ and $\gamma$) should not be constrained in any way.
The fitted classifier can predict the landmine risk from each pattern for each tile, using the coordinate of the tile's center point in the new representation. Thus, each tile has a landmine probability $P_{C_i}$ for each cluster $C_i$. These individual probabilities are then combined as
\begin{equation*}
P_{\mathrm{total}}= 1 - \prod_{i=1}^{m} (1-P_{C_i})
\end{equation*}
to get a total pattern-based landmine probability for a tile.

\subsubsection{Bayesian Prediction} 
The Bayesian Pattern Deminer uses a probabilistic logistic regression model which allows to incorporate prior knowledge into the prediction and provides uncertainty quantification for the prediction \citep{Bishop2006-pl, Hastie2009-ud}. 
At the core of Bayesian logistic regression are prior beliefs about the model parameters, which are expressed as distributions. Three priors are set for the model parameter $\beta \in \mathbb{R}^3$, where $\beta_0$ is the intercept of the model and $\beta_1$ and $\beta_2$ are the coefficients for the $\gamma$ and $\delta$ features. All priors are truncated normal distributions $\mathcal{N}(\beta_i | \mu_i, \sigma_i)$ restricted to a realistic range of values, \emph{i.e.}, ensuring that $\beta_0 \geq 0$ and $\beta_1, \beta_2 < 0$. 
However, the interpretation of the values of $\beta$ is complex, since they refer to the change in log odds \citep{Johnson2022-ie}. Therefore, prior means and standard deviations are computed from six expert estimates. Specifically, an expert is asked to estimate $\gamma_{90}$, $\gamma_{75}$, and $\gamma_{50}$, which are the maximum progresses from a cluster on a pattern with 90\%, 75\%, and 50\% landmine risk, while $\delta=0$. She is also asked to estimate $\delta_{90}$, $\delta_{75}$, and $\delta_{50}$, which are the maximum distances to a cluster pattern with 90\%, 75\%, and 50\% landmine risk while $\gamma=0$. These six estimates are used to derive three equations of the form
\begin{equation*}
\begin{split}
\mathrm{logit}_i = -\beta_0 - \beta_1 \gamma_{i} = -\beta_0 - \beta_2 \delta_{i} \;.
\end{split}
\end{equation*}

The system of equations is solved for three solutions for each parameter ($\beta_0, \beta_1, \beta_2$). The $\mu$ values of the prior distributions are defined as the solution with the median value for $\beta_0$. Furthermore, the $\sigma$ values of the prior distributions are defined as the standard deviation of the three solutions per parameter.
After input from experts in the field of demining, the following set of estimates are used in the experiments to compute the priors for the Bayesian logistic regression model: 
\begin{multicols}{2}
\begin{itemize}
    \item $\gamma_{90}$: 100 meters,
    \item $\gamma_{75}$: 200 meters,
    \item $\gamma_{50}$: 500 meters,
    \item $\delta_{90}$: 50 meters,
    \item $\delta_{75}$: 100 meters,
    \item $\delta_{50}$: 250 meters.
\end{itemize}
\end{multicols}
From the prior and the data likelihood for data $D$, the \textbf{posterior distribution} of the parameter $\beta$ can be derived using the Bayes rule as 
\begin{equation*}
    p(\beta|D) = \frac{p(D|\beta) \; p(\beta)}{p(D)} \;.
\end{equation*}
and the \textbf{predictive distribution} is derived by marginalizing out the posterior distribution as 
\begin{equation*}
    p(y|x) = \int p(y|x, \beta) \; p(\beta|D) \; d\beta \;.
\end{equation*}
Both, the posterior distribution and the predictive distribution, are intractable for Bayesian logistic regression \citep{Bishop2006-pl}. Therefore, sampling techniques are used to approximate these distributions \citep{Murphy2022-nc}. In particular, the \texttt{PyMC} package from \citet{Abril-Pla2023-uv} is used to create posterior distributions based on custom priors and to derive the predictive distribution of landmine risk from each pattern for each tile using Markov Chain Monte Carlo (MCMC) sampling approximations. 
The mean of the sampled predictive distribution is used as the landmine risk for the tile from the pattern. Thus, each tile has a landmine probability $P_{C_i}$ for each cluster $C_i$. These individual probabilities are then combined as
\begin{equation*}
P_{\mathrm{total}}= 1 - \prod_{i=1}^{m} (1-P_{C_i})
\end{equation*}
to get a total pattern-based landmine probability for a tile.

%% file: 1_Content/140_Experiments.tex
\section{Experiments} \label{Chap4}
To answer the research question, the three instances of \input{1_Content/app_name} are first trained with real-world demining data. Afterwards, their performance is evaluated in a virtual landmine clearance environment with real-world demining data and compared to the performance of two baseline deminers to assess the improvement in clearance time efficiency of each instance. 

\subsection{Dataset}
Real-world data from landmine clearance operations in one country is used to train and evaluate the models\footnote{For security and confidentiality reasons, the name of the country and the coordinates of the regions cannot be disclosed. For that reason, actual coordinates of the landmines have been moved to another, arbitrary part of the globe.}. It includes the exact location of the landmines found as geographic coordinates as well as the type, model, and depth of each landmine. 
All landmine locations in the dataset are provided as geographic coordinates, \emph{i.e.}, longitude and latitude. However, in order to apply a metric grid tiling to the region and work with metric distances, the data is first projected to the metric space, in particular, to the coordinate reference system for the region of origin. The software package \texttt{geopandas} from \citet{Jordahl2020-qr} is used to efficiently perform the projections between coordinate reference systems and to work with the data. 

\begin{figure}[H]
    \centering
    \begin{subfigure}[b]{0.45\textwidth}
        \includegraphics[width=\textwidth]{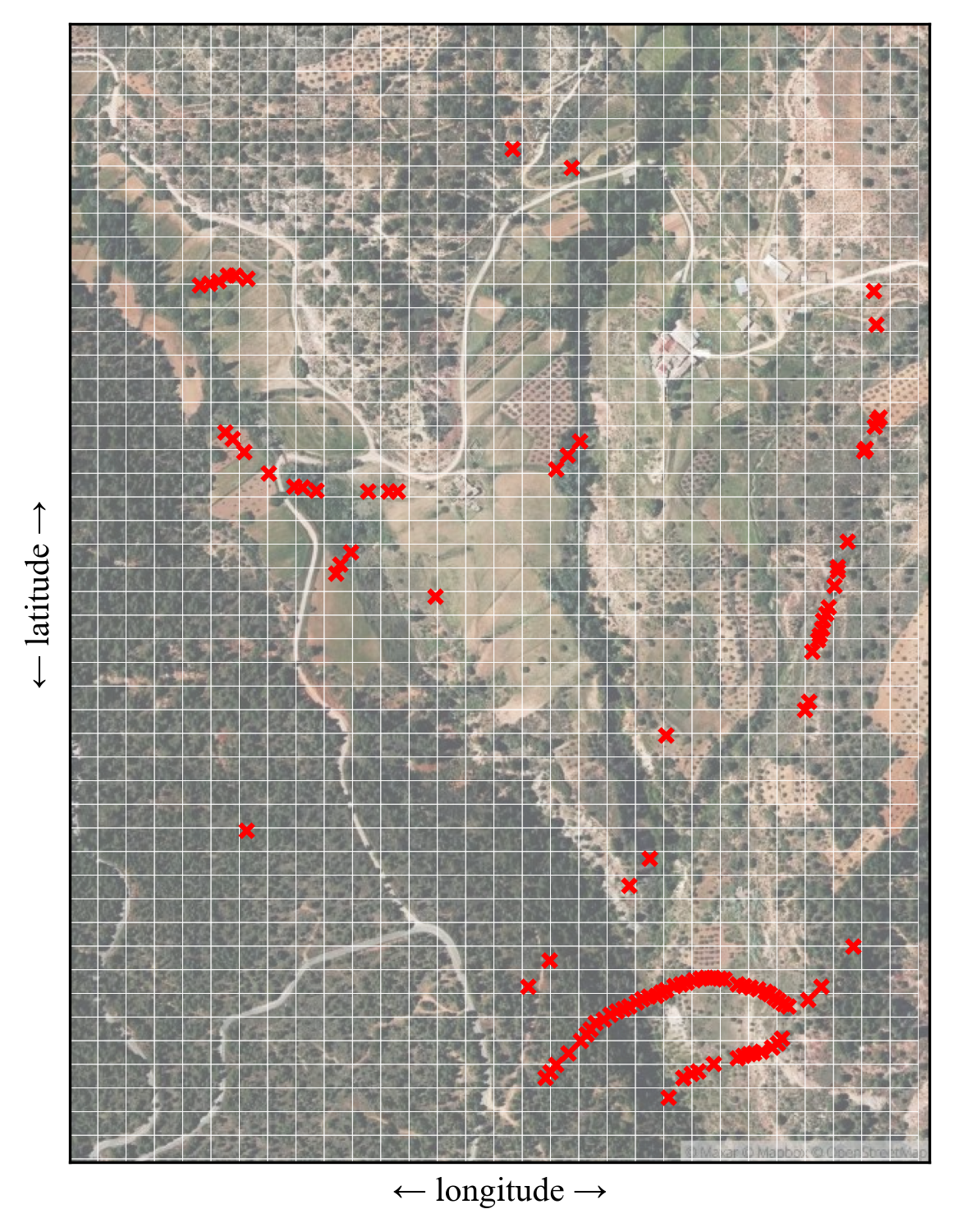}
        \caption{Region "Train 1"}
    \end{subfigure}
    \hfill
    \begin{subfigure}[b]{0.36\textwidth}
        \includegraphics[width=\textwidth]{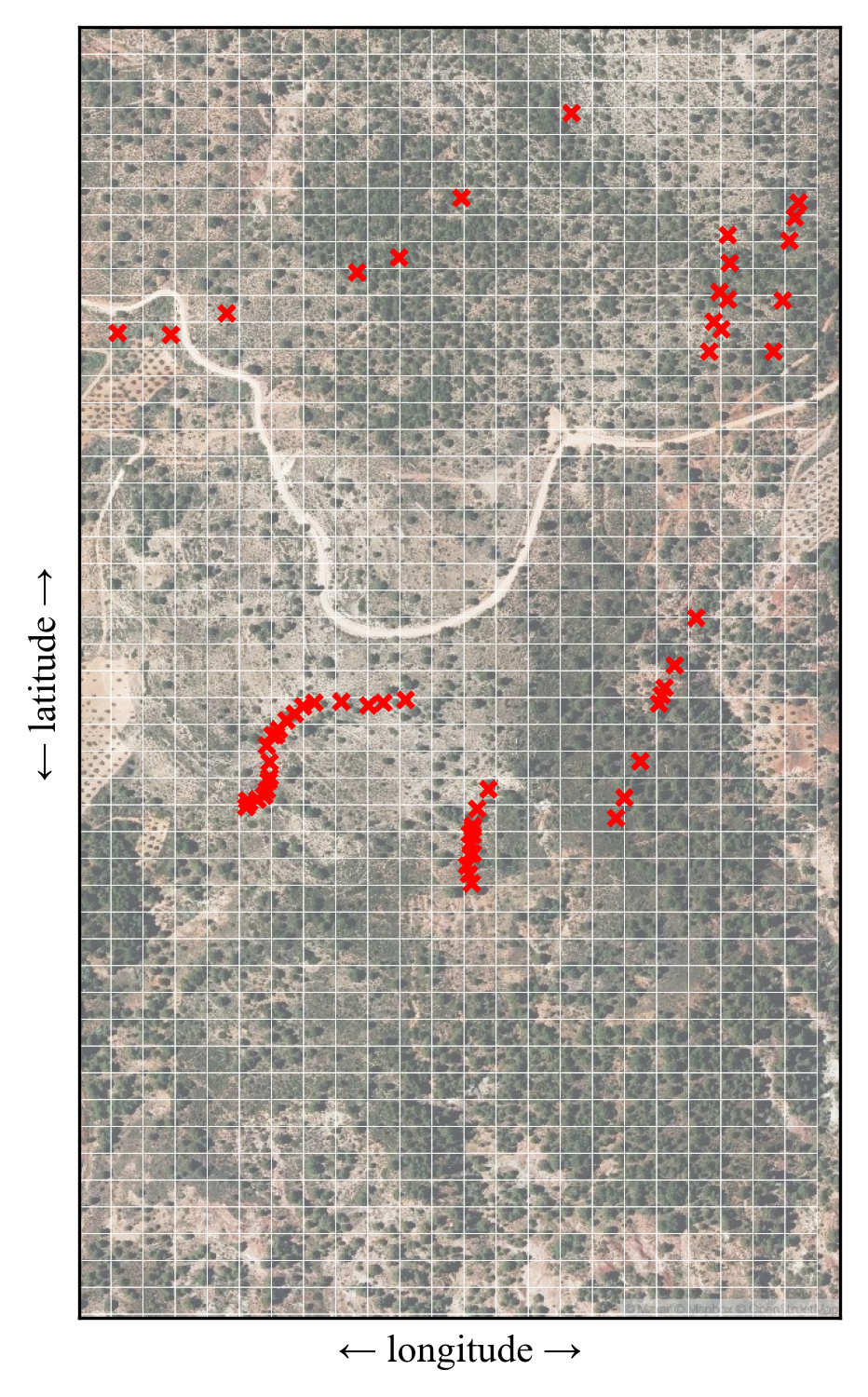}
        \caption{Region "Train 2"}
    \end{subfigure}
    \hfill
    \begin{subfigure}[b]{0.45\textwidth}
        \includegraphics[width=\textwidth]{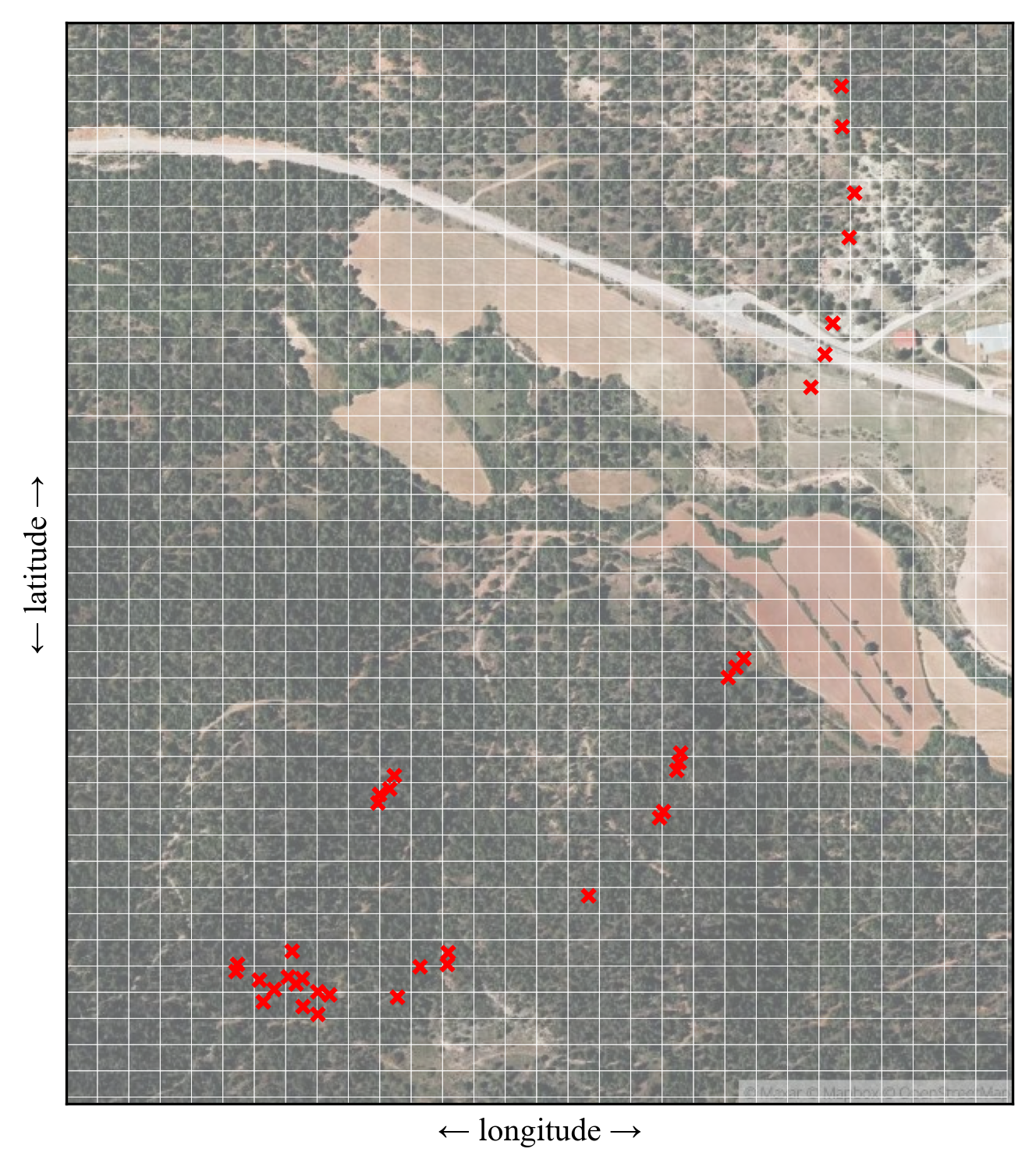}
        \caption{Region "Test 1"}
    \end{subfigure}
    \caption{Regions of interest for training and testing}
    \label{fig:13_Grids}
\end{figure}

To train and test the system, three regions of interest are extracted from the dataset. Two regions are used to train and validate the system, while another region is set aside to test the performance of the system.
A grid with the same tile size of 25 by 25 meters is applied to all regions of interest. 
\Cref{table:13_Regions} lists the characteristics of each region and \Cref{fig:13_Grids} display the regions with their grids and landmine locations.

\begin{center}
\begin{table}[H]
\smaller
\begin{tabularx}{\textwidth}{|XXXX|} 
 \hline
 \textbf{Region} & \textbf{Number of tiles} & \textbf{Number of landmines} & \textbf{Share of tiles with at least one landmine}  \\ 
 \hline
 Train 1 & 1470 & 106 & 4.15 \% \\ 
 \hline
 Train 2 & 1127 & 60 & 3.46 \%  \\
 \hline
 Test 1 & 1260 & 37 & 2.22 \% \\
 \hline
\end{tabularx}
\caption{Train and test regions of interest}
\label{table:13_Regions}
\end{table}
\end{center}

\subsection{Performance Evaluation} \label{Sec_Evaluation}
\begin{figure}[H]
    \begin{center}
        \includegraphics[width=150pt]{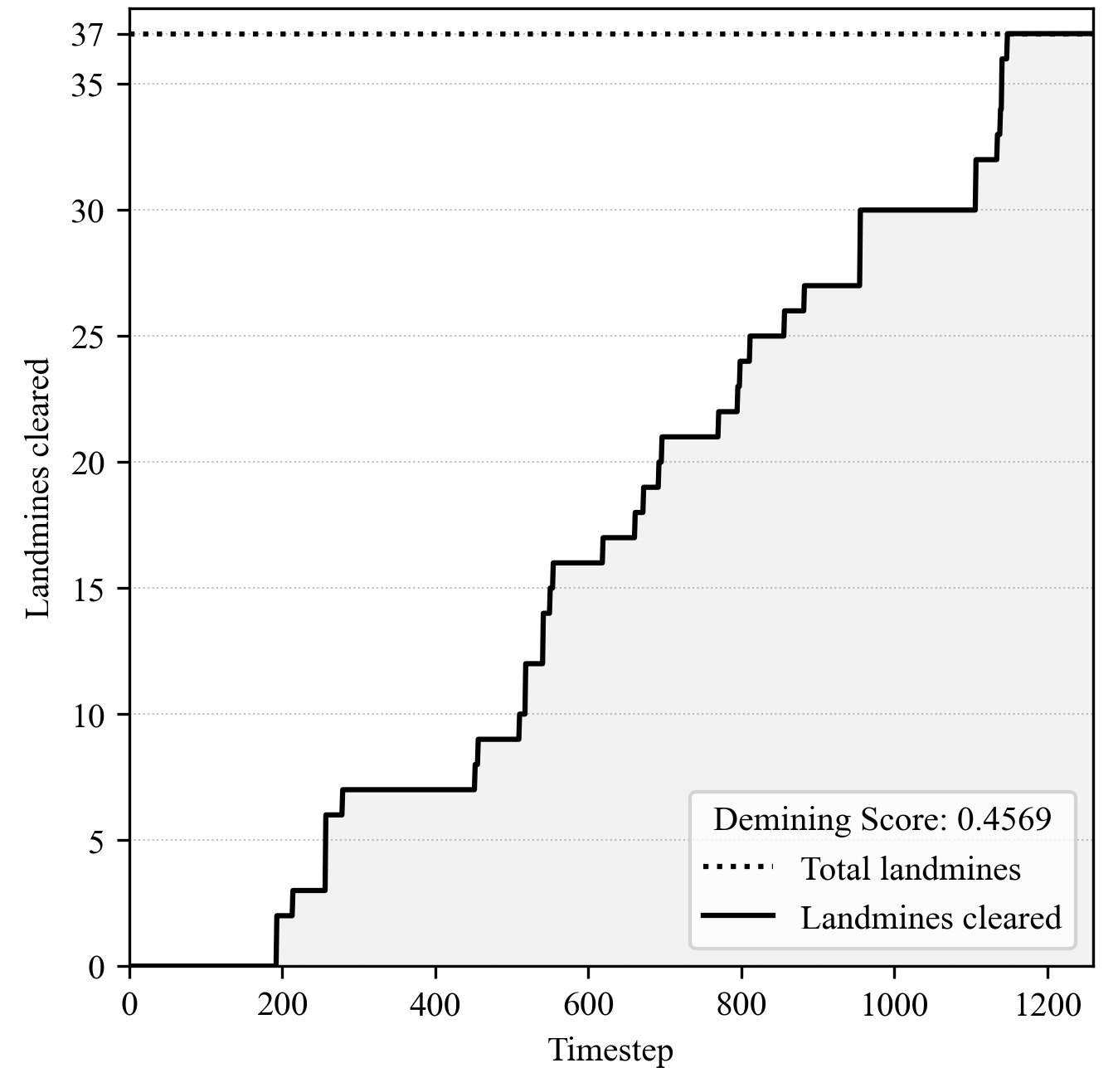}
        \caption{Clearance progress of a random clearance}
        \label{fig:14_Random_Clearance}
    \end{center}
\end{figure}

The clearance performances of all three instances of \input{1_Content/app_name} are evaluated in a virtual landmine clearance environment based on real-world demining data. A virtual deminer is created to clear the test region along a custom route which is based on a landmine risk estimation that is updated online.
Finally, the virtual deminer calculates a \textbf{Demining Score}, which measures how many landmines are cleared and how fast, to evaluate the clearance performance of an instance. 
In addition to the Demining Score, a plot of the clearance processes including the number of landmines found per timestep supports the performance evaluation. \Cref{fig:14_Random_Clearance} shows an exemplary demining plot of a random clearance. 
Finally, the time required to clear 50\%, 75\%, 90\%, and 100\% of landmines is reported as $T_{50}$, $T_{75}$, $T_{90}$, $T_{100}$ for each instance of \input{1_Content/app_name} and the baseline deminers. The $T_{x}$ scores are calculated as the percentage of tiles cleared to find $x$\% of all landmines in the region.

\subsubsection{Virtual Deminer}
\begin{figure}[H]
    \begin{center}
        \includegraphics[width=\textwidth]{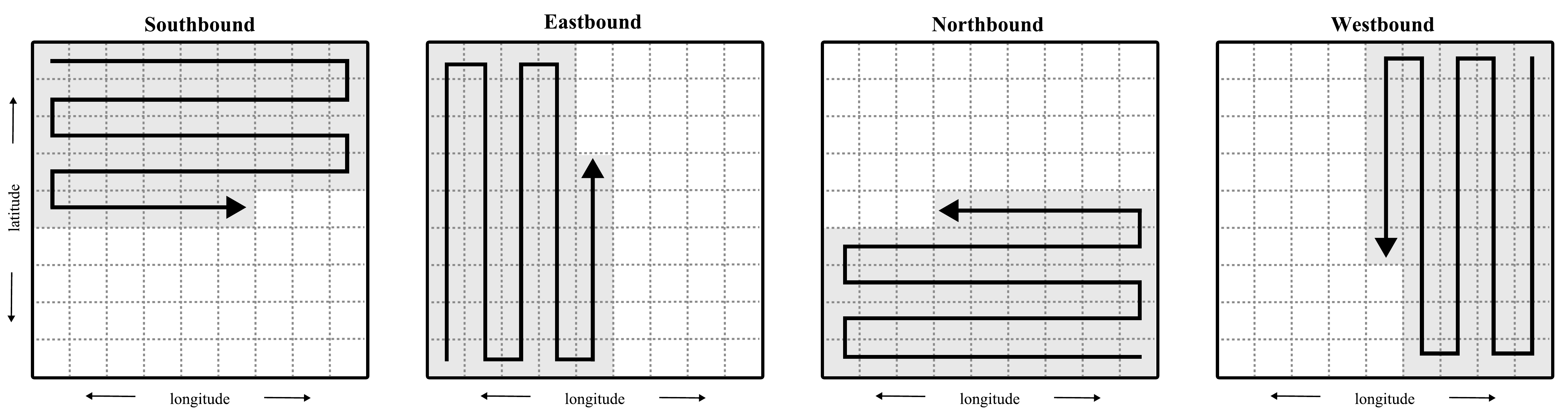}
        \caption{Clearance routes of the Sequential Deminer }
        \label{fig:14_Sequential_Routes}
    \end{center}
\end{figure}

The virtual deminer class implements the necessary functionality to clear a region of interest. 
At each timestep, one reachable tile is selected and \emph{cleared} (\emph{i.e.}, the landmines found in the tile are tracked) until all tiles are cleared. Only tiles that are at the border of the region of interest or for which one of the eight surrounding tiles has already been cleared is reachable in each step. Thus, the virtual deminer is forced to move forward along a path of cleared tiles.  
The selection of tiles can be determined by a simple heuristic (\emph{e.g.}, random or top-down) or by selecting the tile with the highest landmine risk based on an instance's risk estimate. 
The virtual deminers for the three instances of \input{1_Content/app_name} all start with sequential clearance, i.e., row by row from alternating sides, until a first pattern is identified and a landmine risk is calculated for each uncleared tile. A recalculation of the landmine risk is required after a certain number of timesteps, since new patterns may have been revealed in the meantime. To ensure efficient execution, the recalculation is performed every 25 timesteps, \emph{i.e.}, after 25 more tiles have been cleared, for testing. For cross-validation, the recalculation is only performed every 50 timesteps due to the computational complexity of the hyperparameter tuning. The virtual deminer repeats the clearance process four times, starting from all possible sides of the region of interest (see \Cref{fig:14_Sequential_Routes}).

\subsubsection{Demining Score}
The evaluation of an AI system for landmines risk prediction should be aligned with real-world landmine clearance key performance indicators (KPI) to ensure tangible real-world impact. Therefore, the \textbf{Demining Score} is calculated by the virtual deminer to evaluate the three instances of \input{1_Content/app_name} and the baseline deminers. The Demining Score measures the time it takes a deminer to find landmines in a region of interest when following the suggestions of the instance. Therefore, the Demining Score is directly linked to the KPI "deminer days spent per item of explosive ordnance found", which was proposed by \citet{Harutyunyan2023-qc} as one of 11 KPIs for landmine clearance. In addition, the Demining Score can be indirectly linked to the KPIs "cost of square metre of land cleared" and "cost per item of explosive ordnance found", taking into account a reduction in costs if clearance is accelerated \citep{Harutyunyan2023-qc}. 
Specifically, the Demining Score measures the average share of landmines found over the entire clearance process. For example, a deminer that randomly selects tiles to clear will on average achieve a Demining Score slightly below 0.5. An optimal deminer that finds all landmines by clearing only the first few tiles will achieve a Demining Score of (close to) one. 
The score is calculated for a region of interest with a grid of $n$ tiles by tracking the percentage of total landmines found $l_i \in [0, 1]$ for each timestep $i$ of the clearance process. 
The Demining Score $d$ is then calculated as 
\begin{equation*}
d = \frac{1}{n}\sum_{i=1}^{n} l_i  \in [0, 1].
\end{equation*}

\subsubsection{Baselines}
The performances of the three instances of \input{1_Content/app_name} in the virtual demining environment are compared to two baselines.
First, a Random Deminer is used as the simplest baseline. This deminer averages the clearance performance of ten random clearance runs, that clear the virtual environment by randomly selecting the next uncleared tile. 
Second, a Sequential Deminer is used as baseline to model a simple, real-world clearance operation that has no indication of landmine risk and does not adjust its strategy during the clearance process. It averages the clearance performance of four sequential clearance runs, each clearing the region of interest in a different direction, southbound, eastbound, northbound or westbound (see \Cref{fig:14_Sequential_Routes}). 
However, both baselines are likely to perform slightly worse than current real-world clearance operations because they neglect simple heuristics to improve performance. For example, they do not skip tiles that are completely covered by a lake. 

\subsection{Training and Cross Validation} 
The three instances of \input{1_Content/app_name} are trained and optimized using two-fold cross validation. Specifically, each instance is trained and validated for each hyperparameter combination (see \Cref{table:13_Hyperparameters}) on two folds.
First, the respective pattern predictor of the instance is trained on train region 1. Then a complete clearance process is performed on train region 2, including ongoing pattern-risk prediction, and its Demining Score is tracked. 
Afterwards, the process is repeated with training on train region 2 and validation on train region 1. 
A weighted average of the Demining Scores is computed based on the number of tiles in each validation region. The hyperparameters corresponding to the highest weighted Demining Score are selected for each instance. Finally, all three instances are retrained with their optimal hyperparameters on both training regions. The optimal hyperparameters after cross validation are listed in \Cref{table:13_CV_results}.

The differences in optimal \emph{landmine\_weight} between the deminers working with linear and curved patterns suggests that a higher weighting is required for curved pattern prediction. 
Furthermore, the differences between the Curved and the Bayesian Pattern Deminers for \emph{cluster\_max\_distance} and \emph{pc\_smoothness\_factor} show the implications of the expert priors on the optimal hyperparameters for the deminers.

\begin{center}
\begin{table}[H]
\smaller
\begin{tabularx}{\textwidth}{|llX|} 
 \hline
 \textbf{Hyperparameter} & \textbf{Values} & \textbf{Description}  \\ [0.5ex] 
 \hline
 landmine\_weight & 30, 60, 90 & Weighting factor for landmine coordinates  \\
 \hline
 cluster\_max\_distance & 50, 75, 100 & Maximal distance to the cluster for a landmine to be assigned to the cluster (in meter) \\
 \hline
 pc\_smoothness\_factor & 5, 10, 25 & Maximal distance of the principal curve to each landmine of the cluster (in meter)  \\
 \hline
\end{tabularx}
\caption{Hyperparameters for cross validation}
\label{table:13_Hyperparameters}
\end{table}
\end{center}

\begin{center}
\begin{table}[H]
\smaller
\begin{tabularx}{\textwidth}{|Xlll|} 
 \hline
 \textbf{Hyperparameter} & \textbf{Linear Pattern Deminer} & \textbf{Curved Pattern Deminer} & \textbf{Bayesian Pattern Deminer} \\ [0.5ex] 
 \hline
 landmine\_weight & 30 & 90 & 90  \\
 \hline
 cluster\_max\_distance & 100 & 50 & 75 \\
 \hline
 pc\_smoothness\_factor & - & 5 & 25  \\
 \hline
\end{tabularx}
\caption{Optimal hyperparameters after cross validation}
\label{table:13_CV_results}
\end{table}
\end{center}

%% file: 1_Content/150_Results.tex
\section{Results} \label{Chap5}

\newcommand{\cpwidth}{1.0} 

The Demining Scores and $T_x$ values for both baseline deminers and the three instances of \input{1_Content/app_name} are shown in \Cref{table:13_Demining_scores_Connected}.
All instances of \input{1_Content/app_name} significantly outperform both baselines in all metrics. 
Interestingly, the Linear Pattern Deminer performs best of all instances across all metrics except for the $T_{50}$ score, for which the Bayesian Pattern Deminer performs best. 
This means that the Bayesian Pattern Deminer is fastest in clearing 50\% of the landmines in the test region while the Linear Pattern Deminer is fastest in clearing 75\%, 90\%, and 100\% of the landmines. 
Also, this instance has the highest Demining Score, thus, achieves the highest average number of landmines found in each timestep. 
Its Demining Score is 0.1437 points higher than the score of the best baseline deminer, the sequential deminer, meaning that the Linear Pattern Deminer has found 14.37 percentage points more landmines than the best baseline on average. 
Furthermore, its Demining Score is 0.0512 points higher than the score of the Curved Pattern Deminer and  0.0409 points higher than the score of the Bayesian Pattern Deminer, meaning that the Linear Pattern Deminer has on average found about 4 to 5 percentage points more landmines than the other pattern-based deminers.

\begin{center}
\begin{table}[h]
\smaller
\begin{tabularx}{\textwidth}{|lZZZZZ|} 
 \hline
 \textbf{Deminer} & \textbf{Demining Score} & \textbf{$T_{50}$} & \textbf{$T_{75}$} & \textbf{$T_{90}$} & \textbf{$T_{100}$} \\ 
 \hline
 \textbf{Baseline: Random Deminer} & 0.4303 & 55.79\% & 76.35\% & 88.25\% & 99.37\% \\ 
 \hline
 \textbf{Baseline: Sequential Deminer} & 0.5004 & 45.95\% & 73.73\% & 86.11\% & 94.76\% \\ 
 \hline
 \textbf{Linear Pattern Deminer} & \textbf{0.6441} & 39.05\% & \textbf{43.97\%} & \textbf{59.52\%} & \textbf{71.59\%}  \\
 \hline
 \textbf{Curved Pattern Deminer} & 0.5929 & 41.59\% & 53.17\% & 70.08\% & 84.60\%  \\
 \hline
 \textbf{Bayesian Pattern Deminer} & 0.6032 & \textbf{38.10\%} & 50.79\% & 71.43\% & 94.37\%  \\
 \hline
\end{tabularx}
\caption{Clearance performance results}
\label{table:13_Demining_scores_Connected}
\end{table}
\end{center}

\begin{figure}[h]
    \centering
    \includegraphics[width=0.4\textwidth]{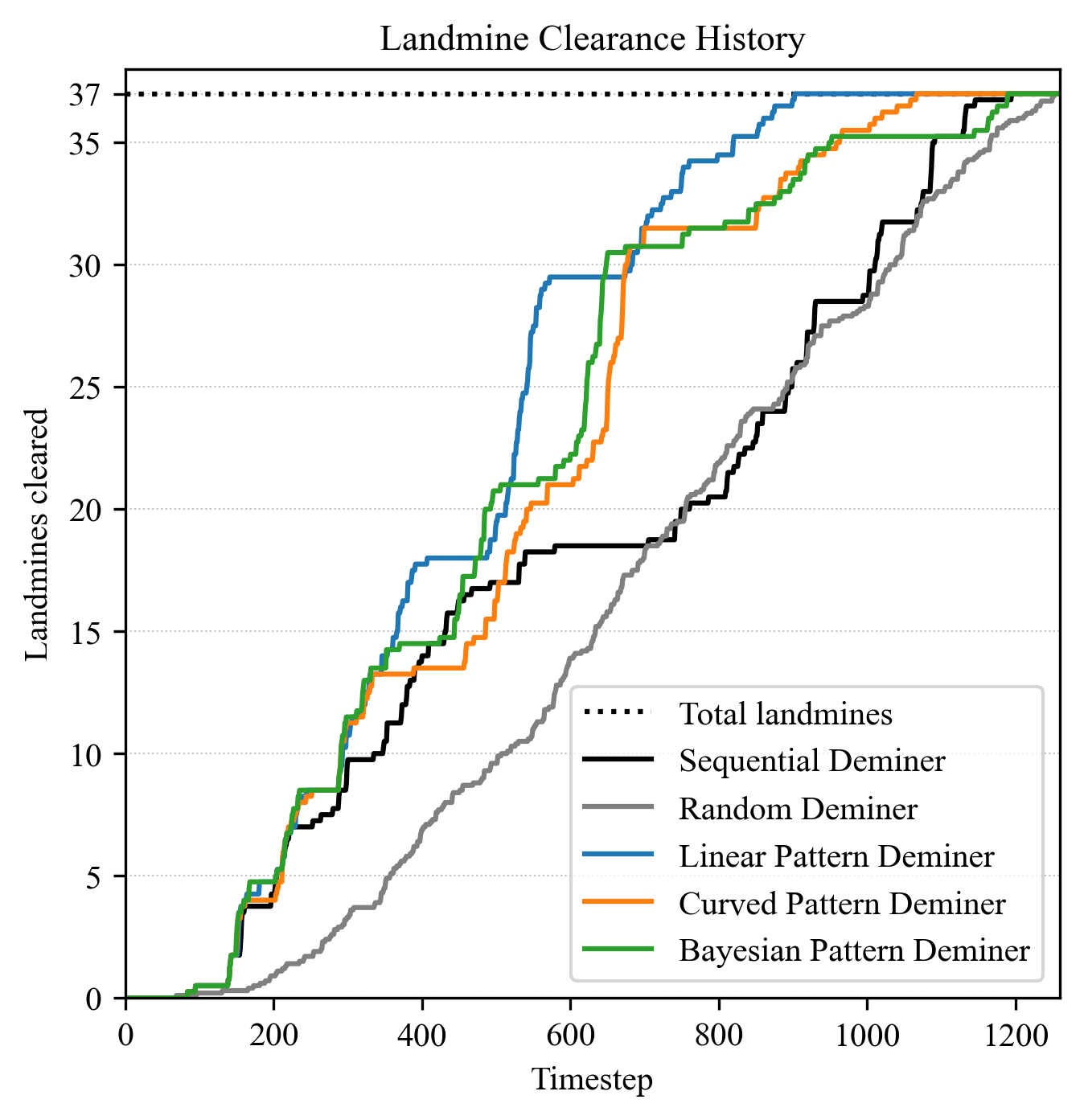}
    \caption{Comparison of clearance progress}
    \label{fig:14_Clearance_Comparison_Connected}
\end{figure}

\Cref{fig:14_Clearance_Comparison_Connected} provides more detailed insights into the clearance performance by showing the clearance history over all timesteps for both baselines and the three instances of \input{1_Content/app_name}. The graphic reveals that all instances and the sequential baseline deminer start of very similarly until timestep 300. However, after this timestep, all instances of \input{1_Content/app_name} start outperforming the sequential deminer because all three instances identify landmine patterns and adjust their clearance routes based on the pattern-based risk prediction. 
In particular, the three instances of \input{1_Content/app_name} achieve significantly higher clearance performance after timestep 600, \emph{i.e.}, after 50\% of the area is cleared. Comparing the clearance progresses of the three instances of \input{1_Content/app_name} shows that all follow a similar trajectory but the Linear Pattern Deminer finds new batches of landmines earlier than the other two instances. 

\subsection{Linear Pattern Deminer}

The Linear Pattern Deminer significantly outperforms both baseline deminers across all metrics. 
Its Demining Score is 0.1437 points higher than the score of the best baseline deminer, the Sequential Deminer, meaning that the Linear Pattern Deminer has found 14.37 percentage points more landmines on average. 
In addition, it is faster than the baseline deminers in clearing 50\%, 75\%, 90\%, and 100\% of the landmines. It requires 24.45\% less time to find all landmines than the fastest baseline deminer.

Furthermore, it outperforms the other two instances of \input{1_Content/app_name} across all metrics except for the $T_{50}$ score, for which the Bayesian Pattern Deminer performs best. 
Its Demining Score is 0.0512 points higher than the score of the Curved Pattern Deminer and  0.0409 points higher than the score of the Bayesian Pattern Deminer, meaning that the Linear Pattern Deminer has on average found about 4 to 5 percentage points more landmines than the other pattern-based deminers.
In addition, it requires 15.38\% less time to find all landmines than the second best pattern-based deminer.

\begin{figure}[H]
        \centering        
        \includegraphics[width=\cpwidth\textwidth]{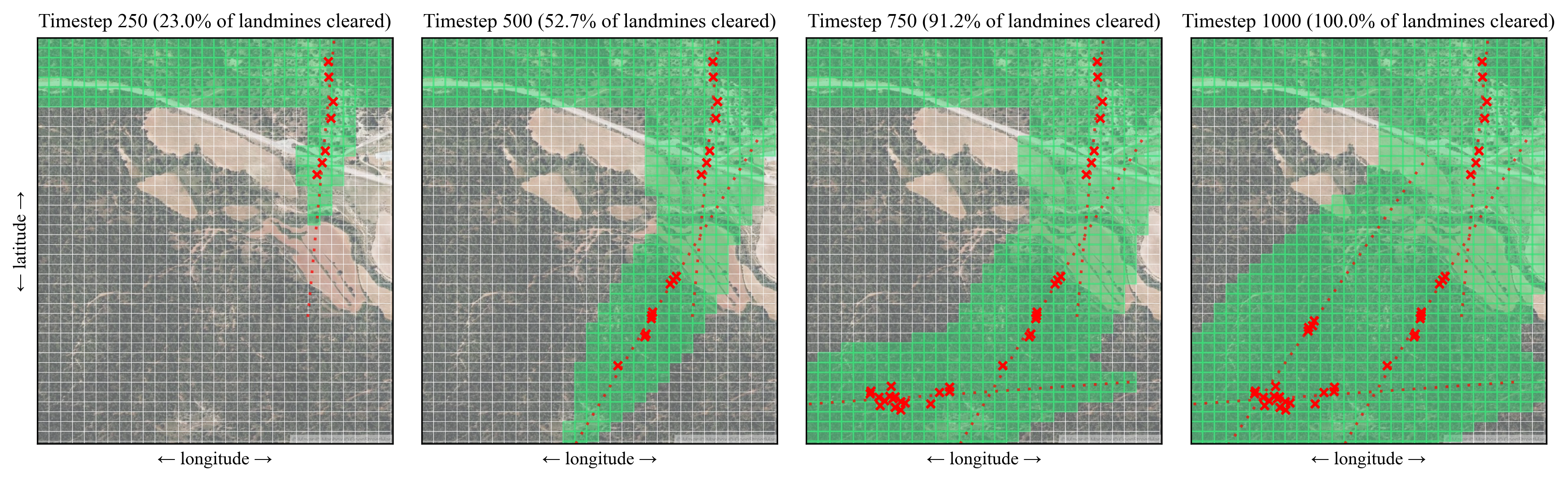}
        \caption{Clearance progress of the Linear Pattern Deminer}
        \label{fig:14_Linear Pattern Deminer Connected_ClearanceImages}
\end{figure}

The landmine patterns found by the Linear Pattern Deminer are shown in \Cref{fig:14_Linear Pattern Deminer Connected_ClearanceImages}. In particular,it shows that the linear patterns mostly cover the landmine clusters and sometimes help to find other clusters. Thus, the linear pattern prediction seems not only to find all the landmines of one cluster, but also to suggest a reasonable search direction to find other clusters that sometimes appear on or near the linear pattern.

\subsection{Curved Pattern Deminer}

The Curved Pattern Deminer significantly outperforms both baseline deminers across all metrics. 
Its Demining Score is 0.0925 points higher than the score of the best baseline deminer, the Sequential Deminer, meaning that the Curved Pattern Deminer has found 9.25 percentage points more landmines on average. 
In addition, it is faster than the baseline deminers in clearing 50\%, 75\%, 90\%, and 100\% of the landmines. It requires 10.72\% less time to find all landmines than the fastest baseline deminer.

However, it performs worst in comparison with the other two instances of \input{1_Content/app_name}. Both achieve better scores across all metrics except for the Bayesian Pattern Deminer which achieves a worse $T_{90}$ and $T_{100}$ score. 
Its Demining Score is 0.0512 points lower than the score of the Linear Pattern Deminer, meaning that it has found about 5 percentage points less landmines than the Linear Pattern Deminer on average.

\begin{figure}[H]
        \centering
        \includegraphics[width=\cpwidth\textwidth]{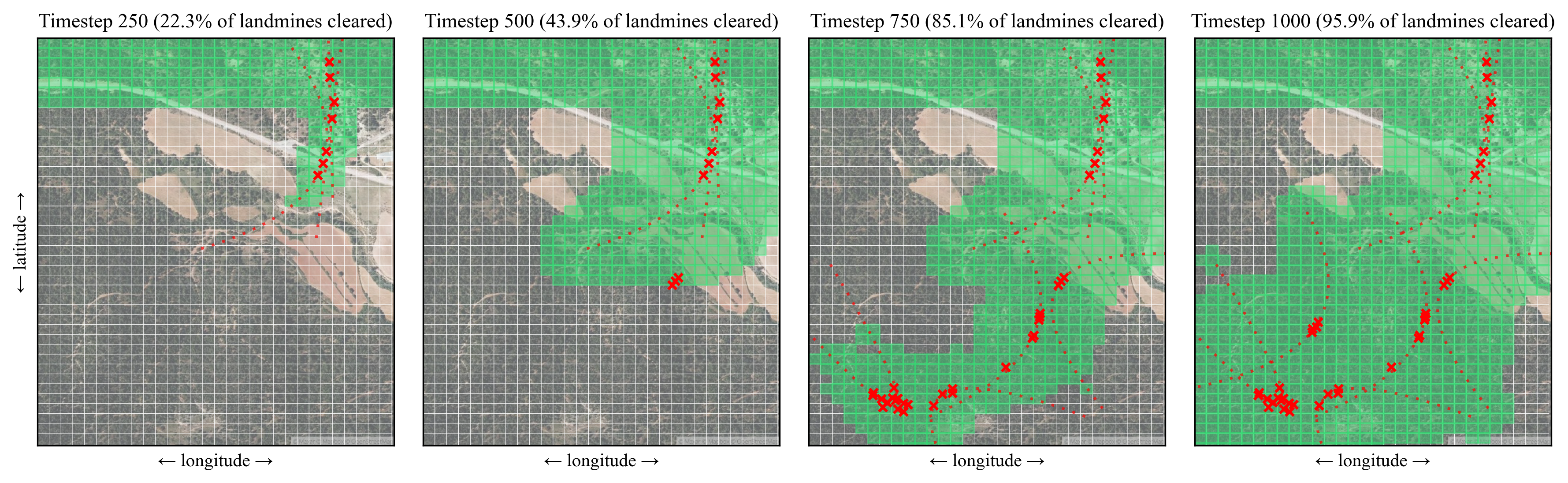}
        \caption{Clearance progress of the Curved Pattern Deminer}
        \label{fig:14_Curved Pattern Deminer Connected_ClearanceImages}
\end{figure}

The landmine patterns found by the Curved Pattern Deminer are shown in \Cref{fig:14_Curved Pattern Deminer Connected_ClearanceImages}.
It shows that the deminer finds patterns that mostly cover the landmine clusters. However, extending these patterns can only sometimes help to find other clusters, because some patterns are predicted as extremely tight curves. Such tight curves suggest a poor search direction that does not point to other clusters, thus degrading clearance efficiency.

\subsection{Bayesian Pattern Deminer}

The Bayesian Pattern Deminer significantly outperforms both baseline deminers across all metrics. 
Its Demining Score is 0.1028 points higher than the score of the best baseline deminer, the Sequential Deminer, meaning that the Bayesian Pattern Deminer has found 10.28 percentage points more landmines on average. 
In addition, it is faster than the baseline deminers in clearing 50\%, 75\%, 90\%, and 100\% of the landmines. 
It requires 17.08\% less time to find the first  50\% of landmines than the fastest baseline deminer. 
However, it requires only 0.41\% less time to find all landmines than the best baseline deminer, thus, achieves only little improvement in time for full clearance of the region of interest. 

In comparison with the other two instances of \input{1_Content/app_name}, the Bayesian Pattern Deminer performs best in finding the first 50\% of landmines, for which it requires 2.43\% less time than the second best pattern-based deminer. 
Yet, its performance degrades over the progress of clearance and it performs worst in finding 90\% and 100\% of landmines, requiring 31.82\% more time than the best patter-based deminer to find all landmines. 
Overall, it achieves the second best Demining Score which is only 0.0103 points higher than the score of the Curved Pattern Deminer. 

\begin{figure}[H]
        \centering
        \includegraphics[width=\cpwidth\textwidth]{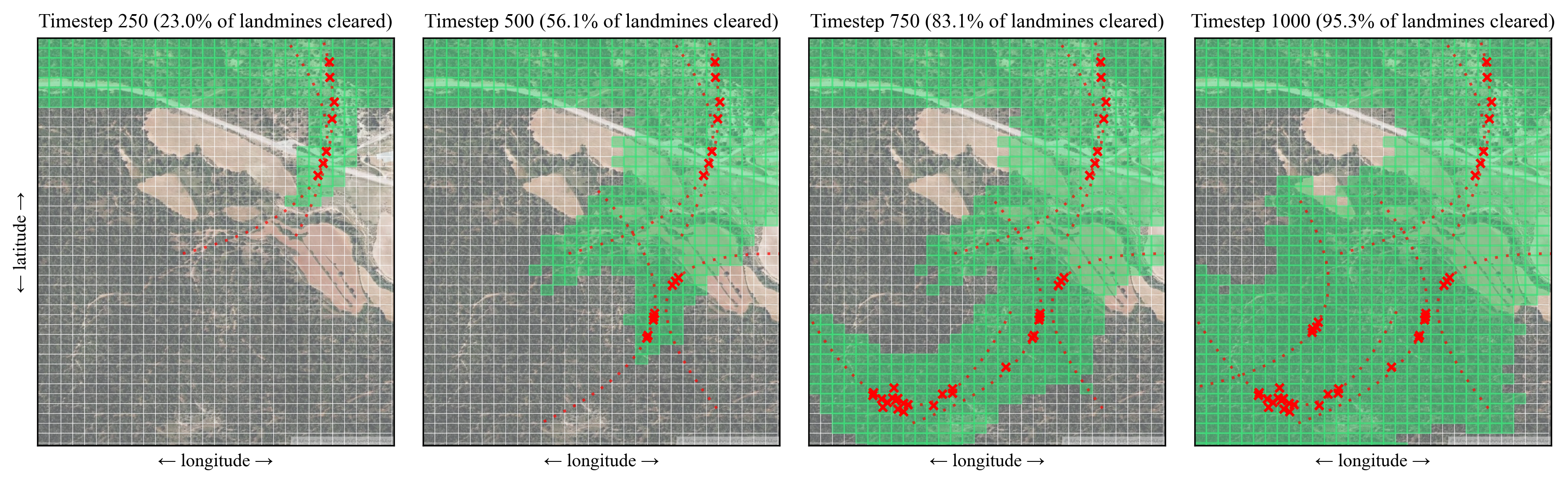}
        \caption{Clearance progress of the Bayesian Pattern Deminer}
        \label{fig:14_Bayesian Pattern Deminer Connected_ClearanceImages}
\end{figure}

The landmine patterns found by the Bayesian Pattern Deminer are shown in \Cref{fig:14_Bayesian Pattern Deminer Connected_ClearanceImages}.
It shows that the deminer finds similar patterns as the Curved Pattern Deminer (see \Cref{fig:14_Curved Pattern Deminer Connected_ClearanceImages}). However, a comparison of the images at timestep 500 also reveals that the priors on pattern-based landmine risk of the Bayesian Pattern Deminer help the instance to not overestimate the risk from one pattern and more quickly focus on new areas in the region of interest. 

%% file: 1_Content/160_Discussion.tex
\section{Discussion} \label{Chap6}

\subsection{Implications for the Research Question}

The results of the three pattern-based instances of \input{1_Content/app_name} provide a positive answer to the research question by introducing approaches that significantly improve clearance time efficiency.  
Specifically, these efficiency improvements are reflected in the higher Demining Scores of the three pattern-based \input{1_Content/app_name} instances compared to the baseline deminers. 
They achieve Demining Scores that are between 0.0925 to 0.1437 points higher than the score of the best baseline, meaning that the average share of landmines found is 9.25 to 14.37 percentage points higher when applying a pattern-based deminer.
Interestingly, the experiments show that the pattern-based deminers can already achieve significant clearance time efficiency improvements early in the clearance process. All three pattern-based instances of \input{1_Content/app_name} significantly outperform the baseline deminers in finding the first 50\% of landmines, requiring between 9.49\% to 17.08\% less time than the best baseline deminer.
However, the conclusion is limited by the performance results of finding all landmines ($T_{100}$ score). While the Linear and Curved Pattern Deminers significantly outperform the baseline deminers, the Bayesian Pattern Deminer does not achieve significant efficiency gains. Thus, its ability to improve clearance time efficiency cannot be confirmed. 

In general, comparing the performance results of the three pattern-based instances of \input{1_Content/app_name} shows that curved patterns do not improve clearance time efficiency over linear patterns. 
This result is surprising because many landmine patterns in the dataset are indeed curved. A possible explanation for the lack of performance improvement could be that the Curved Pattern Deminer predicts some patterns as unrealistically tight curves.
Similarly, the Bayesian interpretation of \input{1_Content/app_name}, which incorporates prior expert knowledge, is unable to achieve significant improvements in clearance time efficiency. It could become more relevant and achieve higher clearance efficiency improvements when \input{1_Content/app_name} is applied to other regions and conflicts that differ more from the training data.
Overall, the results suggest the use of the simpler and more efficient Linear Pattern Deminer. However, further research with additional test data is needed to substantiate this conclusion.

\subsection{Limitations and Future Work}
Although the experiments answer the main research question, they come with limitations that constrain the findings and highlight the need for further research. These limitations stem from the experimental setup and the technical implementation. 

\subsubsection{Limitations from Experimental Setup} The experimental setup introduces significant limitations to the results due to the small amount of data, the limited diversity of the data, and the virtual performance assessment. 

First, the amount of data available for the study is limited. The dataset includes only three regions that contain sufficient landmines for meaningful training. In total, these three regions contain 203 landmines. This limited amount of data may lead to overfitting of the models. In addition, the cross-validation for hyperparameter tuning is not very robust because it can only be run with two folds. Finally, the data allows only one region to be used for testing, which limits the robustness of the performance evaluation. 

Second, the available data lacks diversity. All the data comes from one region in one country, and all the landmines come from one conflict. All landmines in the data are AP landmines of five different types, which are typically planted by hand or from a moving vehicle. Therefore, the system is likely to overfit to such types of landmines and laying techniques.  

Third, the performance evaluation based on a virtual deminer introduces limitations to the results. The approach assumes that each tile requires the same amount of time to clear and does not take into account differences in clearance time due to, for example, difficult terrain \citep{Geneva-International-Centre-for-Humanitarian-Demining2023-om}. Furthermore, the two baseline deminers used in the experiments cannot fully reflect current real-world clearance performance, but are likely to perform worse because they neglect simple heuristics to improve performance. Overall, the performance evaluation approach can only approximate the actual impact on clearance operations. Future work should establish a more accurate performance evaluation by comparing the real-world clearance performance of teams that use \input{1_Content/app_name} and teams that do not.

\subsubsection{Limitations from Technical Implementation} The technical implementation also introduces limitations to the results. These limitations stem from each step in the prediction process; the clustering, the pattern calculation and the pattern risk prediction. 

First, the clustering technique imposes limitations on the results. The implemented DBSCAN clustering algorithm works with a fixed $\epsilon$ value (\textit{cluster\_max\_distance}), which could instead be derived from expert knowledge or other landmine datasets. In addition, other density-based clustering techniques, such as those specified by \citet{Bhattacharjee2021-jm}, could be explored.

Second, there are limitations to the PCA and principal curve pattern calculation. 
In its current implementation, the system approximates all distances to a principal curve with multiple anchor points. Further research could be conducted to find a faster and more accurate approach to this distance computation. Furthermore, sequential principal curves, as introduced by \citet{Li2021-pu}, could be implemented. These allow an existing principal curve to be extended along new landmines during the clearance process, thus reducing the overall computational complexity of fitting the principal curves. 
The results also show that the current implementation of principal curves can lead to unrealistically tight curves that hinder prediction performance (see for example \Cref{fig:14_Curved Pattern Deminer Connected_ClearanceImages}). Future work should explore how to penalize the tightness of a principal curve to ensure realistic patterns. 
Finally, the current implementation is limited to linear and curved patterns, and cannot detect other patterns, such as round clusters resulting from bulk-dispersed landmines, \emph{e.g.}, from an airplane. These would require a different approach to pattern identification and risk prediction. 

Third, the discussed risk prediction has limitations. 
The current implementation uses only (Bayesian) logistic regression and does not test other types of ML models, such as SVMs. In particular, more complex kernel-based models, \emph{e.g.} Gaussian processes, are not analyzed. These could be an interesting extension because of their ability to capture more complex patterns, such as periodic patterns \citep{Bishop2006-pl}. 
Finally, the Bayesian interpretation of pattern prediction has some limitations. In particular, setting the priors can be a difficult task for an expert. The presented method attempts to overcome this challenge by deriving the priors from six expert estimates for landmine risk. However, \citet{Tversky1974-jr} find that such estimates can be flawed due to the heuristics used by experts, which can lead to significant errors.

\subsection{Impact on Clearance Operations}
Beyond the academic scope of this work, it contributes to solving the real-world problem of low efficiency in landmine clearance. 
Although \input{1_Content/app_name} is not yet an end-to-end user-facing information system, the results suggest a novel approach to AI-enabled landmine clearance. Therefore, the results recommend the development of a prototype system that, as anticipated by the \citet{Geneva-International-Centre-for-Humanitarian-Demining2023-wi}, can be a useful tool to support clearance operations and improve their time efficiency. 
In particular, two future implementations of \input{1_Content/app_name} are realistic. 
First, it could be integrated with existing clearance information systems, such as the IMSMA Core offered by the \citet{Geneva-International-Centre-for-Humanitarian-Demining2014-vg}, to support standard clearance operations that are mainly performed by human deminers.
Second, it could be incorporated into modern approaches to UAV-based clearance technologies, such as those developed by \citet{Baur2021-fs}. In such an implementation, \input{1_Content/app_name} could be used, for example, to improve UAV search routes.

This work introduces three main technical innovations to support the development of such AI systems for landmine risk prediction. 
First, \input{1_Content/app_name} implements the first pipeline for iterative, AI-enabled landmine risk prediction, which allows to update the risk prediction after each new landmine is found. 
Second, it implements a Bayesian interpretation of pattern-based landmine risk prediction, which allows bridging the gap between demining expert knowledge and ML. 
Third, it implements a virtual clearance environment and the Demining Score to enable performance evaluation and comparison of AI-enabled clearance systems, which could contribute to an aligned evaluation standard in landmine clearance. 

\subsection{Ethical Considerations}
This research is directly related to armed conflict, peace, and public health. Therefore, the ethical implications of the AI system must be considered. In particular, the three ethical principles of fairness, explainability, and robustness, which are highly relevant to the AI system in question, will be analyzed. These ethical principles are recommended by the \citet{UNESCO2022-dx} and the \citet{European-Parliament2024-xn}.

\input{1_Content/app_name} could inherit significant biases that arise from the underlying data source and affect the fairness of its predictions. However, its architecture allows for high transparency and explainability of the predictions. Finally, the robustness of the system is identified as a major ethical challenge and requires careful mitigation, in particular, by avoiding any public access to the system.

\subsubsection{Fairness} First, the fairness of the system's predictions is an important ethical consideration. Predictions can directly affect people's health if, for example, an area containing landmines is not cleared due to the system's predictions. Therefore, any bias against certain groups of people or regions can have significant consequences. Such bias can come from the training data or from the system design. 
In particular, training data are prone to bias because they are based on geospatial landmine data from previous clearance missions. \citet{Hajjaj2024-lz} note that such data can be biased for several reasons, including security or privacy concerns, cultural and diversity issues, communication barriers, logistical challenges, and other social or economic fluctuations. It may also be biased against regions that lack the resources to support clearance missions \citep{Geneva-International-Centre-for-Humanitarian-Demining2024-ke}. Lack of clearance in such regions results in limited data availability and will propagate the bias to \input{1_Content/app_name}, ultimately causing harm to people living in such regions.
At the time of writing, there is insufficient information about the data collection process to perform a detailed fairness analysis of the data set. However, this analysis remains an important open task for future work. 
Furthermore, the design of the system may inherit biases by optimizing for the limited examples in the training and test data. For example, it is biased towards landmines found in straight and curved patterns but does not consider other patterns such as round clusters. Thus, the design of the system may prevent fair predictions for different groups affected by different types of landmines and landmine patterns. 

\subsubsection{Transparency and Explainability} Second, the transparency and explainability of \input{1_Content/app_name} are important ethical considerations because they allow deminers in the field to understand, interpret, and ultimately overrule the predictions. This could be important if the system makes unrealistic predictions. 
The system's architecture relies on elementary ML techniques that allow to understand and explain its predictions.  
In particular, logistic regression for pattern risk prediction allows interpretation of predictions with coefficient weights \citep{Dwivedi2023-yr}. Furthermore, clustering, PCA and principal curves are unsupervised ML techniques that work with the online clearance data. Therefore, the results of these steps can be illustrated and explained, for example, with the plots featured in \Cref{Chap5}.

\subsubsection{Robustness} Third, the robustness of \input{1_Content/app_name} is an important ethical consideration because adversarial attacks and misuse are likely to occur in armed conflicts and can cause tremendous harm. For example, an adversary in an ongoing conflict could use the system to disrupt clearance missions by fooling the system or learning adversarial landmine patterns. 
The current version of the system has several limitations in its robustness. It is not able to detect any kind of data poisoning, but relies entirely on the given training data. Future versions should compare new training data with trusted data to highlight anomalies. 
Furthermore, the system has not been tested on adversarial examples, which could lead to extremely low or high landmine risk predictions. Future versions should be analyzed for and trained on such adversarial examples to increase the robustness of the system. 
In addition, the system should not be publicly exposed or accessible during training or inference, as model stealing would allow adversaries to use the predictions to place landmines in locations unlikely to be identified by the system. 
Finally, a released version of the system should be accompanied by information security policies and guidelines for its use, as recommended in the International Mine Action Standards on Information Management published by \citet{United-Nations-Mine-Action-Service2023-nv}.

%% file: 1_Content/170_Conclusion.tex
\section{Conclusion} \label{Chap7}

This work introduces \input{1_Content/app_name}, an AI system for risk estimation of remaining explosives. It is the first multi-step ML model that allows for a pattern-based landmine risk prediction. Although the presented implementation of \input{1_Content/app_name} is not yet a user-facing information system, its results suggest the development of a prototype that can be useful to landmine clearance organizations.

The results of the experiments answer the research question by reporting a significant improvement in clearance time efficiency compared to baseline approaches. 
In particular, the results show that linear patterns outperform more complex curved patterns for landmine risk prediction, suggesting the use of the simpler and more efficient Linear Pattern Deminer. 
Finally, the Bayesian interpretation allows the use of prior knowledge for prediction, thereby, introducing an approach that bridges the gap between demining expert knowledge and ML.
Overall, the significant efficiency improvements suggest a novel, AI-based approach to support real-world landmine clearance operations that has the potential to improve clearance KPIs by reducing the time and cost of finding landmines. 

In addition, several directions for future research are revealed. 
On the one hand, future research should test the discussed system with additional, more diverse datasets from real-world clearance operations to confirm performance results and improve generalization. 
On the other hand, future research should extend and improve the proposed technical implementation. In particular, different ML techniques, such as SVMs or Gaussian processes, could be analyzed for the predictors and existing topology prediction models could be integrated into the system. Furthermore, the curved pattern identification with principal curves could be extended to use sequential principal curves and penalize unrealistic patterns. 
Finally, the Bayesian interpretation of \input{1_Content/app_name} requires further refinement through collaboration with demining experts to improve performance and usability. The present paper paves the way to a more systematic, AI-based EO clearance approach that improves on scarce resources to save lives.

%% file: 1_Content/190_Acknowledgements.tex
We would like to thank \textbf{Gregory Cathcart} from the Anti-Personnel Mine Ban Convention Implementation Support Unit for supporting the project with access to real-world clearance data and demining experts. 

%% file: _main.bbl

\begin{thebibliography}{51}


\ifx \showCODEN    \undefined \def \showCODEN     #1{\unskip}     \fi
\ifx \showDOI      \undefined \def \showDOI       #1{#1}\fi
\ifx \showISBNx    \undefined \def \showISBNx     #1{\unskip}     \fi
\ifx \showISBNxiii \undefined \def \showISBNxiii  #1{\unskip}     \fi
\ifx \showISSN     \undefined \def \showISSN      #1{\unskip}     \fi
\ifx \showLCCN     \undefined \def \showLCCN      #1{\unskip}     \fi
\ifx \shownote     \undefined \def \shownote      #1{#1}          \fi
\ifx \showarticletitle \undefined \def \showarticletitle #1{#1}   \fi
\ifx \showURL      \undefined \def \showURL       {\relax}        \fi
\providecommand\bibfield[2]{#2}
\providecommand\bibinfo[2]{#2}
\providecommand\natexlab[1]{#1}
\providecommand\showeprint[2][]{arXiv:#2}

\bibitem[Abril-Pla et~al\mbox{.}(2023)]%
        {Abril-Pla2023-uv}
\bibfield{author}{\bibinfo{person}{Oriol Abril-Pla}, \bibinfo{person}{Virgile Andreani}, \bibinfo{person}{Colin Carroll}, \bibinfo{person}{Larry Dong}, \bibinfo{person}{Christopher~J Fonnesbeck}, \bibinfo{person}{Maxim Kochurov}, \bibinfo{person}{Ravin Kumar}, \bibinfo{person}{Junpeng Lao}, \bibinfo{person}{Christian~C Luhmann}, \bibinfo{person}{Osvaldo~A Martin}, \bibinfo{person}{Michael Osthege}, \bibinfo{person}{Ricardo Vieira}, \bibinfo{person}{Thomas Wiecki}, {and} \bibinfo{person}{Robert Zinkov}.} \bibinfo{year}{2023}\natexlab{}.
\newblock \showarticletitle{{PyMC}: a modern, and comprehensive probabilistic programming framework in Python}.
\newblock \bibinfo{journal}{\emph{PeerJ. Computer science}}  \bibinfo{volume}{9} (\bibinfo{year}{2023}), \bibinfo{pages}{e1516}.
\newblock
\urldef\tempurl%
\url{https://peerj.com/articles/cs-1516}
\showURL{%
\tempurl}


\bibitem[{ACAPS}(2024)]%
        {ACAPS2024-xy}
\bibfield{author}{\bibinfo{person}{{ACAPS}}.} \bibinfo{year}{2024}\natexlab{}.
\newblock \bibinfo{booktitle}{\emph{Ukraine - Humanitarian implications of mine contamination}}.
\newblock \bibinfo{type}{{T}echnical {R}eport}. \bibinfo{institution}{ACAPS}.
\newblock
\urldef\tempurl%
\url{https://www.acaps.org/fileadmin/Data_Product/Main_media/20240124_ACAPS_thematic_report_Ukraine_Analysis_Hub_Humanitarian_implications_of_mine_contamination_.pdf}
\showURL{%
\tempurl}


\bibitem[Alegria et~al\mbox{.}(2011)]%
        {Alegria2011-uq}
\bibfield{author}{\bibinfo{person}{Aura~Cecilia Alegria}, \bibinfo{person}{Hichem Sahli}, {and} \bibinfo{person}{Esteban Zim\'{a}nyi}.} \bibinfo{year}{2011}\natexlab{}.
\newblock \showarticletitle{Application of density analysis for landmine risk mapping}. In \bibinfo{booktitle}{\emph{{IEEE} International Conference on Spatial Data Mining and Geographical Knowledge Services, {ICSDM} 2011, Fuzhou, China, June 29 - July 1, 2011}}. \bibinfo{publisher}{IEEE}, \bibinfo{pages}{223--228}.
\newblock
\urldef\tempurl%
\url{https://doi.org/10.1109/ICSDM.2011.5969036}
\showURL{%
\tempurl}


\bibitem[{Anti-Personnel Mine Ban Convention}(2019)]%
        {Anti-Personnel-Mine-Ban-Convention2019-zy}
\bibfield{author}{\bibinfo{person}{{Anti-Personnel Mine Ban Convention}}.} \bibinfo{year}{2019}\natexlab{}.
\newblock \bibinfo{booktitle}{\emph{Oslo Action Plan}}.
\newblock \bibinfo{type}{{T}echnical {R}eport}. \bibinfo{institution}{Anti-Personnel Mine Ban Convention}.
\newblock
\urldef\tempurl%
\url{https://www.osloreviewconference.org/fileadmin/APMBC-RC4/Fourth-Review-Conference/Oslo-action-plan-en.pdf}
\showURL{%
\tempurl}


\bibitem[Bajic(2010)]%
        {Bajic2010-mu}
\bibfield{author}{\bibinfo{person}{Milan Bajic}.} \bibinfo{year}{2010}\natexlab{}.
\newblock \showarticletitle{The Advanced Intelligence Decision Support System for the Assessment of Mine-suspected Areas}.
\newblock \bibinfo{journal}{\emph{The Journal of Conventional Weapons Destruction}} \bibinfo{volume}{14}, \bibinfo{number}{3} (\bibinfo{year}{2010}), \bibinfo{pages}{28}.
\newblock
\urldef\tempurl%
\url{https://commons.lib.jmu.edu/cisr-journal/vol14/iss3/28}
\showURL{%
\tempurl}


\bibitem[Baur et~al\mbox{.}(2021)]%
        {Baur2021-fs}
\bibfield{author}{\bibinfo{person}{Jasper Baur}, \bibinfo{person}{Gabriel Steinberg}, \bibinfo{person}{Alex Nikulin}, \bibinfo{person}{Kenneth Chiu}, {and} \bibinfo{person}{Timothy de Smet}.} \bibinfo{year}{2021}\natexlab{}.
\newblock \showarticletitle{How to Implement Drones and Machine Learning to Reduce Time, Costs, and Dangers Associated with Landmine Detection}.
\newblock \bibinfo{journal}{\emph{The Journal of Conventional Weapons Destruction}} \bibinfo{volume}{25}, \bibinfo{number}{1} (\bibinfo{year}{2021}), \bibinfo{pages}{29}.
\newblock
\urldef\tempurl%
\url{https://commons.lib.jmu.edu/cisr-journal/vol25/iss1/29}
\showURL{%
\tempurl}


\bibitem[Bhattacharjee and Mitra(2021)]%
        {Bhattacharjee2021-jm}
\bibfield{author}{\bibinfo{person}{Panthadeep Bhattacharjee} {and} \bibinfo{person}{Pinaki Mitra}.} \bibinfo{year}{2021}\natexlab{}.
\newblock \showarticletitle{A survey of density based clustering algorithms}.
\newblock \bibinfo{journal}{\emph{Frontiers of Computer Science}} \bibinfo{volume}{15}, \bibinfo{number}{1} (\bibinfo{year}{2021}).
\newblock
\urldef\tempurl%
\url{https://link.springer.com/article/10.1007/s11704-019-9059-3}
\showURL{%
\tempurl}


\bibitem[Bishop(2006)]%
        {Bishop2006-pl}
\bibfield{author}{\bibinfo{person}{Christopher~M Bishop}.} \bibinfo{year}{2006}\natexlab{}.
\newblock \bibinfo{booktitle}{\emph{Pattern Recognition and Machine Learning}}.
\newblock \bibinfo{publisher}{Springer New York}.
\newblock
\urldef\tempurl%
\url{https://link.springer.com/book/9780387310732}
\showURL{%
\tempurl}


\bibitem[Cumming-Bruce et~al\mbox{.}(2023)]%
        {Cumming-Bruce2023-ry}
\bibfield{author}{\bibinfo{person}{Nick Cumming-Bruce}, \bibinfo{person}{Rula Daoud}, \bibinfo{person}{Alex Frost}, \bibinfo{person}{Emil Hasanov}, \bibinfo{person}{Lucy Pinches}, \bibinfo{person}{Frances Smith}, \bibinfo{person}{Sandra Velasco-P\'{e}rez}, {and} \bibinfo{person}{Louise Wilson}.} \bibinfo{year}{2023}\natexlab{}.
\newblock \bibinfo{booktitle}{\emph{Clearing The Mines 2023}}.
\newblock \bibinfo{type}{{T}echnical {R}eport}. \bibinfo{institution}{Norwegian People's Aid}.
\newblock
\urldef\tempurl%
\url{https://www.mineactionreview.org/assets/downloads/7721_Clearing_the_Mines_2023.pdf}
\showURL{%
\tempurl}


\bibitem[Dwivedi et~al\mbox{.}(2023)]%
        {Dwivedi2023-yr}
\bibfield{author}{\bibinfo{person}{Rudresh Dwivedi}, \bibinfo{person}{Devam Dave}, \bibinfo{person}{Het Naik}, \bibinfo{person}{Smiti Singhal}, \bibinfo{person}{Rana Omer}, \bibinfo{person}{Pankesh Patel}, \bibinfo{person}{Bin Qian}, \bibinfo{person}{Zhenyu Wen}, \bibinfo{person}{Tejal Shah}, \bibinfo{person}{Graham Morgan}, {and} \bibinfo{person}{Rajiv Ranjan}.} \bibinfo{year}{2023}\natexlab{}.
\newblock \showarticletitle{Explainable {AI} ({XAI}): Core ideas, techniques, and solutions}.
\newblock \bibinfo{journal}{\emph{ACM computing surveys}} \bibinfo{volume}{55}, \bibinfo{number}{9} (\bibinfo{year}{2023}), \bibinfo{pages}{1--33}.
\newblock
\urldef\tempurl%
\url{https://dl.acm.org/doi/10.1145/3561048}
\showURL{%
\tempurl}


\bibitem[Ester et~al\mbox{.}(1996)]%
        {Ester1996-yc}
\bibfield{author}{\bibinfo{person}{M Ester}, \bibinfo{person}{H Kriegel}, \bibinfo{person}{J Sander}, {and} \bibinfo{person}{Xiaowei Xu}.} \bibinfo{year}{1996}\natexlab{}.
\newblock \showarticletitle{A density-based algorithm for discovering clusters in large spatial databases with noise}.
\newblock \bibinfo{journal}{\emph{Knowledge Discovery and Data Mining}} (\bibinfo{year}{1996}), \bibinfo{pages}{226--231}.
\newblock
\urldef\tempurl%
\url{https://dl.acm.org/doi/10.5555/3001460.3001507}
\showURL{%
\tempurl}


\bibitem[{European Parliament} and {Council of the European Union}(2024)]%
        {European-Parliament2024-xn}
\bibfield{author}{\bibinfo{person}{{European Parliament}} {and} \bibinfo{person}{{Council of the European Union}}.} \bibinfo{year}{2024}\natexlab{}.
\newblock \showarticletitle{Regulation ({EU}) 2024/1689 of the European Parliament and of the Council of 13 June 2024 laying down harmonised rules on artificial intelligence and amending Regulations ({EC}) No 300/2008, ({EU}) No 167/2013, ({EU}) No 168/2013, ({EU}) 2018/858, ({EU}) 2018/1139 and ({EU}) 2019/2144 and Directives 2014/90/{EU}, ({EU}) 2016/797 and ({EU}) 2020/1828 (Artificial Intelligence Act)}.
\newblock \bibinfo{journal}{\emph{Official Journal of the European Union}} (\bibinfo{year}{2024}).
\newblock
\urldef\tempurl%
\url{https://eur-lex.europa.eu/legal-content/EN/TXT/PDF/?uri=OJ:L_202401689}
\showURL{%
\tempurl}


\bibitem[{Federal Foreign Office Germany}(2024)]%
        {Federal-Foreign-Office-Germany2024-sl}
\bibfield{author}{\bibinfo{person}{{Federal Foreign Office Germany}}.} \bibinfo{year}{2024}\natexlab{}.
\newblock \bibinfo{booktitle}{\emph{Federal Foreign Office Humanitarian Mine Action Strategy within the framework of the humanitarian assistance of the Federal Government}}.
\newblock \bibinfo{type}{{T}echnical {R}eport}. \bibinfo{institution}{Federal Foreign Office Germany}.
\newblock
\urldef\tempurl%
\url{https://www.auswaertiges-amt.de/blob/2651142/4ab9911c36ab10b444354bdb4a1628c1/minenraeumstrategie2024-data.pdf}
\showURL{%
\tempurl}


\bibitem[{Geneva International Centre for Humanitarian Demining}(2014)]%
        {Geneva-International-Centre-for-Humanitarian-Demining2014-vg}
\bibfield{author}{\bibinfo{person}{{Geneva International Centre for Humanitarian Demining}}.} \bibinfo{year}{2014}\natexlab{}.
\newblock \bibinfo{title}{{IMSMA} Core}.
\newblock
\newblock
\urldef\tempurl%
\url{https://www.gichd.org/our-response/information-management/imsma-core/}
\showURL{%
\tempurl}


\bibitem[{Geneva International Centre for Humanitarian Demining}(2023a)]%
        {Geneva-International-Centre-for-Humanitarian-Demining2023-om}
\bibfield{author}{\bibinfo{person}{{Geneva International Centre for Humanitarian Demining}}.} \bibinfo{year}{2023}\natexlab{a}.
\newblock \bibinfo{booktitle}{\emph{Difficult Terrain In Mine Action}}.
\newblock \bibinfo{type}{{T}echnical {R}eport}. \bibinfo{institution}{Geneva International Centre for Humanitarian Demining}.
\newblock
\urldef\tempurl%
\url{https://www.gichd.org/fileadmin/user_upload/GICHD_Difficult_Terrain_A5_10_WEB.pdf}
\showURL{%
\tempurl}


\bibitem[{Geneva International Centre for Humanitarian Demining}(2023b)]%
        {Geneva-International-Centre-for-Humanitarian-Demining2023-wi}
\bibfield{author}{\bibinfo{person}{{Geneva International Centre for Humanitarian Demining}}.} \bibinfo{year}{2023}\natexlab{b}.
\newblock \bibinfo{booktitle}{\emph{{GICHD} Innovation Conference 2023 Report}}.
\newblock \bibinfo{type}{{T}echnical {R}eport}. \bibinfo{institution}{Geneva International Centre for Humanitarian Demining}.
\newblock
\urldef\tempurl%
\url{https://www.gichd.org/fileadmin/uploads/gichd/Photos/Innovation_Conference_2023/GICHD_Innovation_Conference_Report.pdf}
\showURL{%
\tempurl}


\bibitem[{Geneva International Centre for Humanitarian Demining}(2024)]%
        {Geneva-International-Centre-for-Humanitarian-Demining2024-ya}
\bibfield{author}{\bibinfo{person}{{Geneva International Centre for Humanitarian Demining}}.} \bibinfo{year}{2024}\natexlab{}.
\newblock \bibinfo{booktitle}{\emph{{GICHD} Annual Report 2023}}.
\newblock \bibinfo{type}{{T}echnical {R}eport}. \bibinfo{institution}{Geneva International Centre for Humanitarian Demining}.
\newblock
\urldef\tempurl%
\url{https://www.gichd.org/fileadmin/uploads/gichd/Media/Annual_Report_2023/GICHD_2023_AnnualReport_FINAL.pdf}
\showURL{%
\tempurl}


\bibitem[{Geneva International Centre for Humanitarian Demining} and {Symbio Impact Ltd.}(2024)]%
        {Geneva-International-Centre-for-Humanitarian-Demining2024-ke}
\bibfield{author}{\bibinfo{person}{{Geneva International Centre for Humanitarian Demining}} {and} \bibinfo{person}{{Symbio Impact Ltd.}}} \bibinfo{year}{2024}\natexlab{}.
\newblock \bibinfo{booktitle}{\emph{Innovative Finance for Mine Action: Needs and Potential Solutions}}.
\newblock \bibinfo{type}{{T}echnical {R}eport}. \bibinfo{institution}{{Geneva International Centre for Humanitarian Demining}; {Symbio Impact Ltd.};}.
\newblock
\urldef\tempurl%
\url{https://www.gichd.org/fileadmin/user_upload/INNOVATIVE_FINANCE_FOR_MINE_ACTION_NEEDS_AND_POTENTIAL_SOLUTIONS.pdf}
\showURL{%
\tempurl}


\bibitem[Hajjaj et~al\mbox{.}(2024)]%
        {Hajjaj2024-lz}
\bibfield{author}{\bibinfo{person}{Maysa Hajjaj}, \bibinfo{person}{Lauren Burrows}, \bibinfo{person}{Teia Rogers}, \bibinfo{person}{Natalia Lozano}, \bibinfo{person}{Sarah~Kamal Elias}, {and} \bibinfo{person}{Samban Seng}.} \bibinfo{year}{2024}\natexlab{}.
\newblock \showarticletitle{Inclusive data mangement: Reporting, storing, and sharing of information on beneficiaries in the mine action sector}.
\newblock \bibinfo{journal}{\emph{The Journal of Conventional Weapons Destruction}} \bibinfo{volume}{28}, \bibinfo{number}{1} (\bibinfo{year}{2024}), \bibinfo{pages}{6}.
\newblock
\urldef\tempurl%
\url{https://commons.lib.jmu.edu/cisr-journal/vol28/iss1/6}
\showURL{%
\tempurl}


\bibitem[Harutyunyan et~al\mbox{.}(2023)]%
        {Harutyunyan2023-qc}
\bibfield{author}{\bibinfo{person}{Armen Harutyunyan}, \bibinfo{person}{Danielle Payne}, \bibinfo{person}{David Hewitson}, {and} \bibinfo{person}{Raphaela Lark}.} \bibinfo{year}{2023}\natexlab{}.
\newblock \bibinfo{booktitle}{\emph{Operational Efficiency in Mine Action}}.
\newblock \bibinfo{type}{{T}echnical {R}eport}. \bibinfo{institution}{Geneva International Centre for Humanitarian Demining}.
\newblock
\urldef\tempurl%
\url{https://www.gichd.org/publications-resources/publications/operational-efficiency-in-mine-action/}
\showURL{%
\tempurl}


\bibitem[Hastie and Stuetzle(1989)]%
        {Hastie1989-wt}
\bibfield{author}{\bibinfo{person}{Trevor Hastie} {and} \bibinfo{person}{Werner Stuetzle}.} \bibinfo{year}{1989}\natexlab{}.
\newblock \showarticletitle{Principal Curves}.
\newblock \bibinfo{journal}{\emph{J. Amer. Statist. Assoc.}} \bibinfo{volume}{84}, \bibinfo{number}{406} (\bibinfo{year}{1989}), \bibinfo{pages}{502--516}.
\newblock
\urldef\tempurl%
\url{https://www.tandfonline.com/doi/abs/10.1080/01621459.1989.10478797}
\showURL{%
\tempurl}


\bibitem[Hastie et~al\mbox{.}(2009)]%
        {Hastie2009-ud}
\bibfield{author}{\bibinfo{person}{Trevor Hastie}, \bibinfo{person}{Robert Tibshirani}, {and} \bibinfo{person}{Jerome Friedman}.} \bibinfo{year}{2009}\natexlab{}.
\newblock \bibinfo{booktitle}{\emph{The elements of statistical learning: Data mining, inference, and prediction, second edition} (\bibinfo{edition}{2} ed.)}.
\newblock \bibinfo{publisher}{Springer}.
\newblock
\urldef\tempurl%
\url{https://link.springer.com/book/10.1007/978-0-387-84858-7}
\showURL{%
\tempurl}


\bibitem[Hofmann and Juergensen(2017)]%
        {Hofmann2017-pl}
\bibfield{author}{\bibinfo{person}{Ursign Hofmann} {and} \bibinfo{person}{Olaf Juergensen}.} \bibinfo{year}{2017}\natexlab{}.
\newblock \bibinfo{booktitle}{\emph{Leaving no one Behind: Mine Action and the Sustainable Development Goals}}.
\newblock \bibinfo{type}{{T}echnical {R}eport}. \bibinfo{institution}{Geneva International Centre for Humanitarian Demining (GICHD); United Nations Development Programme (UNDP)}.
\newblock
\urldef\tempurl%
\url{https://www.undp.org/publications/mine-action-and-sustainable-development-goals}
\showURL{%
\tempurl}


\bibitem[{International Campaign to Ban Landmines}(2023)]%
        {International-Campaign-to-Ban-Landmines2023-pc}
\bibfield{author}{\bibinfo{person}{{International Campaign to Ban Landmines}}.} \bibinfo{year}{2023}\natexlab{}.
\newblock \bibinfo{booktitle}{\emph{Landmine Monitor 2023}}.
\newblock \bibinfo{type}{{T}echnical {R}eport}. \bibinfo{institution}{International Campaign to Ban Landmines}.
\newblock
\urldef\tempurl%
\url{https://www.the-monitor.org/en-gb/reports/2023/landmine-monitor-2023.aspx}
\showURL{%
\tempurl}


\bibitem[Johnson et~al\mbox{.}(2022)]%
        {Johnson2022-ie}
\bibfield{author}{\bibinfo{person}{Alicia~A Johnson}, \bibinfo{person}{Miles~Q Ott}, {and} \bibinfo{person}{{Mine Dogucu}}.} \bibinfo{year}{2022}\natexlab{}.
\newblock \bibinfo{booktitle}{\emph{Bayes rules!: An introduction to applied Bayesian modeling} (\bibinfo{edition}{1} ed.)}.
\newblock \bibinfo{publisher}{Chapman \& Hall/CRC}.
\newblock
\urldef\tempurl%
\url{https://www.routledge.com/Bayes-Rules-An-Introduction-to-Applied-Bayesian-Modeling/Johnson-Ott-Dogucu/p/book/9780367255398}
\showURL{%
\tempurl}


\bibitem[Jolliffe(2002)]%
        {Jolliffe2002-hq}
\bibfield{author}{\bibinfo{person}{Ian~T Jolliffe}.} \bibinfo{year}{2002}\natexlab{}.
\newblock \bibinfo{booktitle}{\emph{Principal Component Analysis} (\bibinfo{edition}{2} ed.)}.
\newblock \bibinfo{publisher}{Springer}.
\newblock
\urldef\tempurl%
\url{https://link.springer.com/book/10.1007/b98835}
\showURL{%
\tempurl}


\bibitem[Jordahl et~al\mbox{.}(2020)]%
        {Jordahl2020-qr}
\bibfield{author}{\bibinfo{person}{Kelsey Jordahl}, \bibinfo{person}{Joris Van~den Bossche}, \bibinfo{person}{Martin Fleischmann}, \bibinfo{person}{Jacob Wasserman}, \bibinfo{person}{James McBride}, \bibinfo{person}{Jeffrey Gerard}, \bibinfo{person}{Jeff Tratner}, \bibinfo{person}{Matthew Perry}, \bibinfo{person}{Adrian~Garcia Badaracco}, \bibinfo{person}{Carson Farmer}, \bibinfo{person}{Geir~Arne Hjelle}, \bibinfo{person}{Alan~D Snow}, \bibinfo{person}{Micah Cochran}, \bibinfo{person}{Sean Gillies}, \bibinfo{person}{Lucas Culbertson}, \bibinfo{person}{Matt Bartos}, \bibinfo{person}{Nick Eubank}, \bibinfo{person}{{Maxalbert}}, \bibinfo{person}{Aleksey Bilogur}, \bibinfo{person}{Sergio Rey}, \bibinfo{person}{Christopher Ren}, \bibinfo{person}{Dani Arribas-Bel}, \bibinfo{person}{Leah Wasser}, \bibinfo{person}{Levi~John Wolf}, \bibinfo{person}{Martin Journois}, \bibinfo{person}{Joshua Wilson}, \bibinfo{person}{Adam Greenhall}, \bibinfo{person}{Chris Holdgraf}, \bibinfo{person}{{Filipe}}, {and}
  \bibinfo{person}{Fran\c{c}ois Leblanc}.} \bibinfo{year}{2020}\natexlab{}.
\newblock \bibinfo{title}{geopandas/geopandas: {v0}.8.1}.
\newblock
\newblock
\urldef\tempurl%
\url{https://doi.org/10.5281/zenodo.3946761}
\showURL{%
\tempurl}


\bibitem[Jurafsky and Martin(2008)]%
        {Jurafsky2008-wu}
\bibfield{author}{\bibinfo{person}{Daniel Jurafsky} {and} \bibinfo{person}{James~H Martin}.} \bibinfo{year}{2008}\natexlab{}.
\newblock \bibinfo{booktitle}{\emph{Speech and Language Processing: An Introduction to Natural Language Processing, Computational Linguistics, and Speech Recognition}}.
\newblock \bibinfo{publisher}{Prentice Hall}.
\newblock


\bibitem[Kischelewski et~al\mbox{.}(2025)]%
        {Kischelewski2025-qy}
\bibfield{author}{\bibinfo{person}{Bj{\"{o}}rn Kischelewski}, \bibinfo{person}{Gregory Cathcart}, \bibinfo{person}{David Wahl}, {and} \bibinfo{person}{Benjamin Guedj}.} \bibinfo{year}{2025}\natexlab{}.
\newblock \showarticletitle{{AI} for explosive ordnance detection in clearance operations: The state of research}.
\newblock \bibinfo{journal}{\emph{arXiv [cs.LG]}} (\bibinfo{year}{2025}).
\newblock
\urldef\tempurl%
\url{http://arxiv.org/abs/2411.05813}
\showURL{%
\tempurl}


\bibitem[Lee et~al\mbox{.}(2006)]%
        {Lee2006-qe}
\bibfield{author}{\bibinfo{person}{Yun-Seok Lee}, \bibinfo{person}{Han-Suh Koo}, {and} \bibinfo{person}{Chang-Sung Jeong}.} \bibinfo{year}{2006}\natexlab{}.
\newblock \showarticletitle{A straight line detection using principal component analysis}.
\newblock \bibinfo{journal}{\emph{Pattern recognition letters}} \bibinfo{volume}{27}, \bibinfo{number}{14} (\bibinfo{year}{2006}), \bibinfo{pages}{1744--1754}.
\newblock
\urldef\tempurl%
\url{http://dx.doi.org/10.1016/j.patrec.2006.04.016}
\showURL{%
\tempurl}


\bibitem[Li and Guedj(2021)]%
        {Li2021-pu}
\bibfield{author}{\bibinfo{person}{Le Li} {and} \bibinfo{person}{Benjamin Guedj}.} \bibinfo{year}{2021}\natexlab{}.
\newblock \showarticletitle{Sequential Learning of Principal Curves: Summarizing Data Streams on the Fly}.
\newblock \bibinfo{journal}{\emph{Entropy}} \bibinfo{volume}{23}, \bibinfo{number}{11} (\bibinfo{year}{2021}).
\newblock
\urldef\tempurl%
\url{https://www.mdpi.com/1099-4300/23/11/1534}
\showURL{%
\tempurl}


\bibitem[McCullagh and Nelder(1989)]%
        {McCullagh1989-bx}
\bibfield{author}{\bibinfo{person}{P McCullagh} {and} \bibinfo{person}{J~A Nelder}.} \bibinfo{year}{1989}\natexlab{}.
\newblock \bibinfo{booktitle}{\emph{Generalized Linear Models} (\bibinfo{edition}{2nd edition} ed.)}.
\newblock \bibinfo{publisher}{Routledge}.
\newblock
\urldef\tempurl%
\url{https://api.taylorfrancis.com/content/books/mono/download?identifierName=doi&identifierValue=10.1201/9780203753736&type=googlepdf}
\showURL{%
\tempurl}


\bibitem[{Mine Action Review}(2023)]%
        {Mine-Action-Review2023-it}
\bibfield{author}{\bibinfo{person}{{Mine Action Review}}.} \bibinfo{year}{2023}\natexlab{}.
\newblock \bibinfo{booktitle}{\emph{A Guide To The Oslo Action Plan And Results Of 2023 Monitoring: Survey and Clearance}}.
\newblock \bibinfo{type}{{T}echnical {R}eport}. \bibinfo{institution}{Mine Action Review}.
\newblock
\urldef\tempurl%
\url{https://www.mineactionreview.org/assets/downloads/A_Guide_to_the_Oslo_Action_Plan_and_Results_of_2023_Monitoring.pdf}
\showURL{%
\tempurl}


\bibitem[Murphy(2022)]%
        {Murphy2022-nc}
\bibfield{author}{\bibinfo{person}{Kevin~P Murphy}.} \bibinfo{year}{2022}\natexlab{}.
\newblock \bibinfo{booktitle}{\emph{Probabilistic machine learning: An introduction}}.
\newblock \bibinfo{publisher}{MIT Press}.
\newblock
\urldef\tempurl%
\url{https://books.google.de/books?hl=en&lr=&id=wrZNEAAAQBAJ&oi=fnd&pg=PR27&dq=probabilistic+machine+learning+an+introduction&ots=LaneKzbpWm&sig=3pF14tVM0_KUqSKR2ZGEKdDzg2U}
\showURL{%
\tempurl}


\bibitem[Osmolovska(2023)]%
        {Osmolovska2023-ze}
\bibfield{author}{\bibinfo{person}{Iuliia Osmolovska}.} \bibinfo{year}{2023}\natexlab{}.
\newblock \bibinfo{booktitle}{\emph{Walking on Fire: Demining in Ukraine}}.
\newblock \bibinfo{type}{{T}echnical {R}eport}. \bibinfo{institution}{{GLOBSEC}}.
\newblock
\urldef\tempurl%
\url{https://www.globsecusfoundation.org/publications/walking-on-fire-demining-in-ukraine/}
\showURL{%
\tempurl}


\bibitem[Osmolovska and Bilyk(2024)]%
        {Osmolovska2024-mi}
\bibfield{author}{\bibinfo{person}{Iuliia Osmolovska} {and} \bibinfo{person}{Nataliia Bilyk}.} \bibinfo{year}{2024}\natexlab{}.
\newblock \bibinfo{booktitle}{\emph{Cleaning the Augean Stables: Humanitarian Demining in Ukraine}}.
\newblock \bibinfo{type}{{T}echnical {R}eport}. \bibinfo{institution}{{GLOBSEC}}.
\newblock
\urldef\tempurl%
\url{https://www.globsecusfoundation.org/publications/cleaning-the-augean-stables-humanitarian-demining-in-ukraine/}
\showURL{%
\tempurl}


\bibitem[Pedregosa et~al\mbox{.}(2011)]%
        {Pedregosa2011-nt}
\bibfield{author}{\bibinfo{person}{F Pedregosa}, \bibinfo{person}{G Varoquaux}, \bibinfo{person}{A Gramfort}, \bibinfo{person}{V Michel}, \bibinfo{person}{B Thirion}, \bibinfo{person}{O Grisel}, \bibinfo{person}{M Blondel}, \bibinfo{person}{P Prettenhofer}, \bibinfo{person}{R Weiss}, \bibinfo{person}{V Dubourg}, \bibinfo{person}{J Vanderplas}, \bibinfo{person}{A Passos}, \bibinfo{person}{D Cournapeau}, \bibinfo{person}{M Brucher}, \bibinfo{person}{M Perrot}, {and} \bibinfo{person}{E Duchesnay}.} \bibinfo{year}{2011}\natexlab{}.
\newblock \showarticletitle{Scikit-learn: Machine Learning in Python}.
\newblock \bibinfo{journal}{\emph{Journal of Machine Learning Research}}  \bibinfo{volume}{12} (\bibinfo{year}{2011}), \bibinfo{pages}{2825--2830}.
\newblock


\bibitem[Rafique et~al\mbox{.}(2019)]%
        {Rafique2019-pt}
\bibfield{author}{\bibinfo{person}{Waqas Rafique}, \bibinfo{person}{Dawei Zheng}, \bibinfo{person}{Jamie Barras}, \bibinfo{person}{Sagar Joglekar}, {and} \bibinfo{person}{Panagiotis Kosmas}.} \bibinfo{year}{2019}\natexlab{}.
\newblock \showarticletitle{Predictive Analysis of Landmine Risk}.
\newblock \bibinfo{journal}{\emph{IEEE Access}}  \bibinfo{volume}{7} (\bibinfo{year}{2019}), \bibinfo{pages}{107259--107269}.
\newblock
\urldef\tempurl%
\url{https://ieeexplore.ieee.org/document/8765724}
\showURL{%
\tempurl}


\bibitem[Riese et~al\mbox{.}(2006)]%
        {Riese2006-bm}
\bibfield{author}{\bibinfo{person}{Stephen~R Riese}, \bibinfo{person}{Donald~E Brown}, {and} \bibinfo{person}{Yacov~Y Haimes}.} \bibinfo{year}{2006}\natexlab{}.
\newblock \showarticletitle{Estimating the probability of landmine contamination}.
\newblock \bibinfo{journal}{\emph{Military operations research}} \bibinfo{volume}{11}, \bibinfo{number}{3} (\bibinfo{year}{2006}), \bibinfo{pages}{49--61}.
\newblock
\urldef\tempurl%
\url{http://openurl.ingenta.com/content/xref?genre=article&issn=1082-5983&volume=11&issue=3&spage=49}
\showURL{%
\tempurl}


\bibitem[Rubio et~al\mbox{.}(2023)]%
        {Rubio2023-hy}
\bibfield{author}{\bibinfo{person}{Mateo~Dulce Rubio}, \bibinfo{person}{Siqi Zeng}, \bibinfo{person}{Qi Wang}, \bibinfo{person}{Didier Alvarado}, \bibinfo{person}{Francisco Moreno}, \bibinfo{person}{Hoda Heidari}, {and} \bibinfo{person}{Fei Fang}.} \bibinfo{year}{2023}\natexlab{}.
\newblock \showarticletitle{{RELand}: Risk Estimation of Landmines via Interpretable Invariant Risk Minimization}.
\newblock \bibinfo{journal}{\emph{ACM Journal on Computing and Sustainable Societies}} \bibinfo{volume}{2}, \bibinfo{number}{2} (\bibinfo{year}{2023}).
\newblock
\urldef\tempurl%
\url{https://dl.acm.org/doi/10.1145/3648437}
\showURL{%
\tempurl}


\bibitem[Saliba et~al\mbox{.}(2024)]%
        {Saliba2024-lx}
\bibfield{author}{\bibinfo{person}{Adib Saliba}, \bibinfo{person}{Kifah Tout}, \bibinfo{person}{Chamseddine Zaki}, {and} \bibinfo{person}{Christophe Claramunt}.} \bibinfo{year}{2024}\natexlab{}.
\newblock \showarticletitle{A location-based model using {GIS} with machine learning, and a human-based approach for demining a post-war region}.
\newblock \bibinfo{journal}{\emph{Journal of location based services}} \bibinfo{volume}{18}, \bibinfo{number}{2} (\bibinfo{year}{2024}), \bibinfo{pages}{162--184}.
\newblock
\urldef\tempurl%
\url{https://www.tandfonline.com/doi/full/10.1080/17489725.2023.2298803}
\showURL{%
\tempurl}


\bibitem[Schubert et~al\mbox{.}(2017)]%
        {Schubert2017-pv}
\bibfield{author}{\bibinfo{person}{Erich Schubert}, \bibinfo{person}{J{\"{o}}rg Sander}, \bibinfo{person}{Martin Ester}, \bibinfo{person}{Hans~Peter Kriegel}, {and} \bibinfo{person}{Xiaowei Xu}.} \bibinfo{year}{2017}\natexlab{}.
\newblock \showarticletitle{{DBSCAN} Revisited, Revisited: Why and how you should (still) use {DBSCAN}}.
\newblock \bibinfo{journal}{\emph{ACM transactions on database systems}} \bibinfo{volume}{42}, \bibinfo{number}{3} (\bibinfo{year}{2017}), \bibinfo{pages}{1--21}.
\newblock
\urldef\tempurl%
\url{https://dl.acm.org/doi/10.1145/3068335}
\showURL{%
\tempurl}


\bibitem[Shalev-Shwartz and Ben-David(2014)]%
        {Shalev-Shwartz2014-tc}
\bibfield{author}{\bibinfo{person}{Shai Shalev-Shwartz} {and} \bibinfo{person}{Shai Ben-David}.} \bibinfo{year}{2014}\natexlab{}.
\newblock \bibinfo{booktitle}{\emph{Understanding machine learning: From theory to algorithms}}.
\newblock \bibinfo{publisher}{Cambridge University Press}.
\newblock
\urldef\tempurl%
\url{https://www.cambridge.org/core/books/understanding-machine-learning/3059695661405D25673058E43C8BE2A6}
\showURL{%
\tempurl}


\bibitem[Thomas and Cathcart(2010)]%
        {Thomas2010-vg}
\bibfield{author}{\bibinfo{person}{Alan~M Thomas} {and} \bibinfo{person}{J~Michael Cathcart}.} \bibinfo{year}{2010}\natexlab{}.
\newblock \showarticletitle{Applications of Grid Pattern Matching to the Detection of Buried Landmines}.
\newblock \bibinfo{journal}{\emph{IEEE transactions on geoscience and remote sensing: a publication of the IEEE Geoscience and Remote Sensing Society}} \bibinfo{volume}{48}, \bibinfo{number}{9} (\bibinfo{year}{2010}), \bibinfo{pages}{3465--3470}.
\newblock
\urldef\tempurl%
\url{http://dx.doi.org/10.1109/TGRS.2010.2046740}
\showURL{%
\tempurl}


\bibitem[Toscano(2021)]%
        {Toscano2021-im}
\bibfield{author}{\bibinfo{person}{Stefano Toscano}.} \bibinfo{year}{2021}\natexlab{}.
\newblock \showarticletitle{Past, Present, Future: Mine Action in Motion}.
\newblock \bibinfo{journal}{\emph{The Journal of Conventional Weapons Destruction}} \bibinfo{volume}{25}, \bibinfo{number}{1} (\bibinfo{year}{2021}), \bibinfo{pages}{4}.
\newblock
\urldef\tempurl%
\url{https://commons.lib.jmu.edu/cisr-journal/vol25/iss1/4/}
\showURL{%
\tempurl}


\bibitem[Tversky and Kahneman(1974)]%
        {Tversky1974-jr}
\bibfield{author}{\bibinfo{person}{A Tversky} {and} \bibinfo{person}{D Kahneman}.} \bibinfo{year}{1974}\natexlab{}.
\newblock \showarticletitle{Judgment under uncertainty: Heuristics and biases: Biases in judgments reveal some heuristics of thinking under uncertainty}.
\newblock \bibinfo{journal}{\emph{Science (New York, N.Y.)}} \bibinfo{volume}{185}, \bibinfo{number}{4157} (\bibinfo{year}{1974}), \bibinfo{pages}{1124--1131}.
\newblock
\urldef\tempurl%
\url{https://www.science.org/doi/pdf/10.1126/science.185.4157.1124}
\showURL{%
\tempurl}


\bibitem[{UNESCO}(2022)]%
        {UNESCO2022-dx}
\bibfield{author}{\bibinfo{person}{{UNESCO}}.} \bibinfo{year}{2022}\natexlab{}.
\newblock \bibinfo{booktitle}{\emph{Recommendation on the Ethics of Artificial Intelligence}}.
\newblock \bibinfo{type}{{T}echnical {R}eport}. \bibinfo{institution}{United Nations Educational, Scientific and Cultural Organization}.
\newblock
\urldef\tempurl%
\url{https://www.unesco.org/en/articles/recommendation-ethics-artificial-intelligence}
\showURL{%
\tempurl}


\bibitem[{United Nations}(2023)]%
        {United-Nations2023-zs}
\bibfield{author}{\bibinfo{person}{{United Nations}}.} \bibinfo{year}{2023}\natexlab{}.
\newblock \bibinfo{booktitle}{\emph{The United Nations Mine Action Strategy}}.
\newblock \bibinfo{type}{{T}echnical {R}eport}. \bibinfo{institution}{United Nations}.
\newblock
\urldef\tempurl%
\url{https://www.mineaction.org/sites/default/files/publications/un_mine_action_strategy_2024.pdf}
\showURL{%
\tempurl}


\bibitem[{United Nations General Assembly}(2023)]%
        {United-Nations-General-Assembly2023-wn}
\bibfield{author}{\bibinfo{person}{{United Nations General Assembly}}.} \bibinfo{year}{2023}\natexlab{}.
\newblock \bibinfo{title}{A/{RES}/78/70 - Resolution on Assistance in mine action}.
\newblock
\newblock
\urldef\tempurl%
\url{https://www.unmas.org/sites/default/files/documents/un_ga_res_78-70.pdf}
\showURL{%
\tempurl}


\bibitem[{United Nations Mine Action Service}(2023)]%
        {United-Nations-Mine-Action-Service2023-nv}
\bibfield{author}{\bibinfo{person}{{United Nations Mine Action Service}}.} \bibinfo{year}{2023}\natexlab{}.
\newblock \bibinfo{booktitle}{\emph{{IMAS} 05.10 - Information management in mine action}}.
\newblock \bibinfo{type}{{T}echnical {R}eport}. \bibinfo{institution}{{United Nations Mine Action Service};}.
\newblock
\urldef\tempurl%
\url{https://www.mineactionstandards.org/fileadmin/uploads/imas/Standards/English/IMAS_05.10_Ed.2_Am.2.pdf}
\showURL{%
\tempurl}


\bibitem[Zhang(2020)]%
        {Zhang2020-ks}
\bibfield{author}{\bibinfo{person}{Stephen Zhang}.} \bibinfo{year}{2020}\natexlab{}.
\newblock \bibinfo{title}{pcurvepy: Principal curves implementation (Hastie '89) in Python}.
\newblock
\newblock
\urldef\tempurl%
\url{https://github.com/zsteve/pcurvepy}
\showURL{%
\tempurl}


\end{thebibliography}
